\documentclass[journal,10pt]{IEEEtran}

\usepackage{booktabs} 
\usepackage{graphicx}
\usepackage{amsmath}
\usepackage{amssymb}
\usepackage[caption=false]{subfig}
\usepackage{bbding}
\usepackage{multirow}
\usepackage{makecell}
\usepackage{cite}
\usepackage{overpic}
\usepackage{xcolor}
\usepackage{pifont}
\usepackage{colortbl}
\usepackage{threeparttable}
\usepackage[hidelinks]{hyperref}

\graphicspath{{./Imgs/}{./Imgs/Samples/}{./Imgs/Curve/}{./Imgs/Authors/}}
\DeclareGraphicsExtensions{.pdf,.jpg,.png}

\newcommand{\figref}[1]{Fig.~\ref{#1}}
\newcommand{\tabref}[1]{Table~\ref{#1}}
\newcommand{\secref}[1]{Section~\ref{#1}}

\newcommand{\equref}[1]{Eq. (\ref{#1})}

\def\ie{\textit{i.e.}}
\def\eg{\textit{e.g.}}

\def\etc{\textit{etc}}
\def\etal{\textit{et al.}~}
\def\sArt{{state-of-the-art~}}
\definecolor{mycolor}{rgb}{.92,.92,.92}

\begin{document}
\title{Boosting Salient Object Detection with Transformer-based Asymmetric Bilateral U-Net}

\author{Yu Qiu,
        Yun Liu\href{mailto:VAGRANTLYUN@GMAIL.COM}{\Envelope},
        Le Zhang,
        Haotian Lu,
        Jing Xu\href{mailto:xujing@nankai.edu.cn}{\Envelope}
\thanks{This work was supported in part by Science and Technology Planning Project of Tianjin (17JCZDJC30700 and 18ZXZNGX00310), and in part by Fundamental Research Funds for the Central Universities of Nankai University (63191402).
\textit{(Corresponding authors: Yun Liu; Jing Xu.)}
}
\thanks{Yu Qiu, Haotian Lu, and Jing Xu are with the College of Artificial Intelligence, Nankai University, Tianjin 300350, China.}
\thanks{Yun Liu is with the Institute for Infocomm Research (I2R), Agency for Science, Technology and Research (A*STAR), 138632, Singapore.}
\thanks{Le Zhang is with the School of Information and Communication Engineering, University of Electronic Science and Technology of China (UESTC), Chengdu 611731, China.}}

\markboth{IEEE Transactions on Circuits and Systems for Video Technology}
{Boosting Salient Object Detection with Transformer-based Asymmetric Bilateral U-Net}

\maketitle

\begin{abstract}
Existing salient object detection (SOD) methods mainly rely on U-shaped convolution neural networks (CNNs) with skip connections to combine the global contexts and local spatial details that are crucial for locating salient objects and refining object details, respectively. Despite great successes, the ability of CNNs in learning global contexts is limited. Recently, the vision transformer has achieved revolutionary progress in computer vision owing to its powerful modeling of global dependencies. However, directly applying the transformer to SOD is suboptimal because the transformer lacks the ability to learn local spatial representations. To this end, this paper explores the combination of transformers and CNNs to learn both global and local representations for SOD. We propose a transformer-based Asymmetric Bilateral U-Net (ABiU-Net). The asymmetric bilateral encoder has a transformer path and a lightweight CNN path, where the two paths communicate at each encoder stage to learn complementary global contexts and local spatial details, respectively. The asymmetric bilateral decoder also consists of two paths to process features from the transformer and CNN encoder paths, with communication at each decoder stage for decoding coarse salient object locations and fine-grained object details, respectively. Such communication between the two encoder/decoder paths enables AbiU-Net to learn complementary global and local representations, taking advantage of the natural merits of transformers and CNNs, respectively. Hence, ABiU-Net provides a new perspective for transformer-based SOD. Extensive experiments demonstrate that ABiU-Net performs favorably against previous state-of-the-art SOD methods. The code is available at \url{https://github.com/yuqiuyuqiu/ABiU-Net}.
\end{abstract}


\IEEEpeerreviewmaketitle

\section{Introduction} \label{sec:intro}
\IEEEPARstart{S}{alient} object detection (SOD) aims at detecting the 
most visually conspicuous objects or regions in an image
\cite{cheng2014global,liu2021dna,liu2020lightweight,wang2020deep,tu2020edge,li2021dense,zhang2022progressive,zhang2022tcrnet,mei2021exploring}.
It has a wide range of computer vision applications such as 
human-robot interaction \cite{meger2008curious},
content-aware image editing \cite{cheng2010repfinder},
image retrieval \cite{gao2013visual},
object recognition \cite{ren2013region},
image thumbnailing \cite{marchesotti2009framework},
weakly supervised learning \cite{liu2020leveraging}, \etc.
In the last decade, convolutional neural networks (CNNs)
have significantly pushed forward this field.
Intuitively, the global contextual information (existing in the top 
CNN layers) is essential for \textit{locating salient objects}, 
while the local fine-grained information (existing in the bottom 
CNN layers) is helpful in \textit{refining object details}
\cite{cheng2014global,luo2017non,liu2018picanet,hu2020sac,zhang2021engaging,zhang2022tcrnet,mei2021exploring}.
This is why the U-shaped encoder-decoder CNNs have dominated this field \cite{wang2015deep,lee2016deep,wang2016saliency,li2016deep,zhang2017amulet,zhang2017learning,luo2017non,wang2017stagewise,liu2018picanet,li2018contour,chen2018reverse,liu2020lightweight,liu2021samnet,qiu2019revisiting,qiu2020simple,wu2019cascaded,zhao2019egnet,liu2021dna,qiu2021miniseg,liu2022poolnet+,sun2021ampnet},
where the encoder extracts multi-level deep features from 
raw images and the decoder integrates the extracted features 
with skip connections to make image-to-image predictions
\cite{wang2015deep,lee2016deep,wang2016saliency,li2016deep,zhang2017amulet,zhang2017learning,luo2017non,wang2017stagewise,liu2018picanet,li2018contour,chen2018reverse,liu2020lightweight,liu2021samnet,zhang2022dhnet}.
The encoder is usually the existing CNN backbones,
\eg, ResNet \cite{he2016deep},
while most efforts are put into the design of the decoder
\cite{qiu2019revisiting,qiu2020simple,wu2019cascaded,liu2022poolnet+,zhao2019egnet}.
Although remarkable progress has been seen in this direction,
CNN-based encoders share the intrinsic limitation of extracting 
features from images in a local manner.
The lack of powerful global modeling has been the main bottleneck for CNN-based SOD.

To this end, we note that recent popular transformer networks
\cite{vaswani2017attention,dosovitskiy2021image}
provide a new perspective on this problem.
Originating from machine translation \cite{vaswani2017attention}, 
transformers entirely reply on self-attention to model global 
dependencies of sequence data directly.
Viewing image patches as tokens (words) in natural language 
processing (NLP) applications \cite{dosovitskiy2021image}, 
transformers can be applied to learn powerful global feature 
representations for images.

Since the global relationship modeling of transformers is 
beneficial to SOD for locating salient objects in a natural scene,
some works have tried to bring transformers into SOD
\cite{liu2021visual,mao2021transformer}.
However, we note that existing transformer-based SOD methods
\cite{liu2021visual,mao2021transformer} 
entirely rely on the transformer to extract global features 
by using the transformer as the encoder.
They ignore the effect of local representations,
which is also essential for SOD in refining object details
\cite{zhang2017amulet,zhang2017learning,luo2017non,wang2017stagewise,liu2018picanet,li2018contour,chen2018reverse,liu2020lightweight,liu2021samnet,qiu2019revisiting,qiu2020simple,wu2019cascaded,zhao2019egnet,liu2022poolnet+,zhang2021engaging,zhang2022tcrnet}, 
as mentioned above.
Therefore, existing SOD methods have gone from one extreme to the other, \ie, from the lack of powerful global modeling (CNN-based methods) to the lack of local representation learning (transformer-based methods).

\begin{figure*}[!tb]
    \centering
    \includegraphics[width=.95\linewidth]{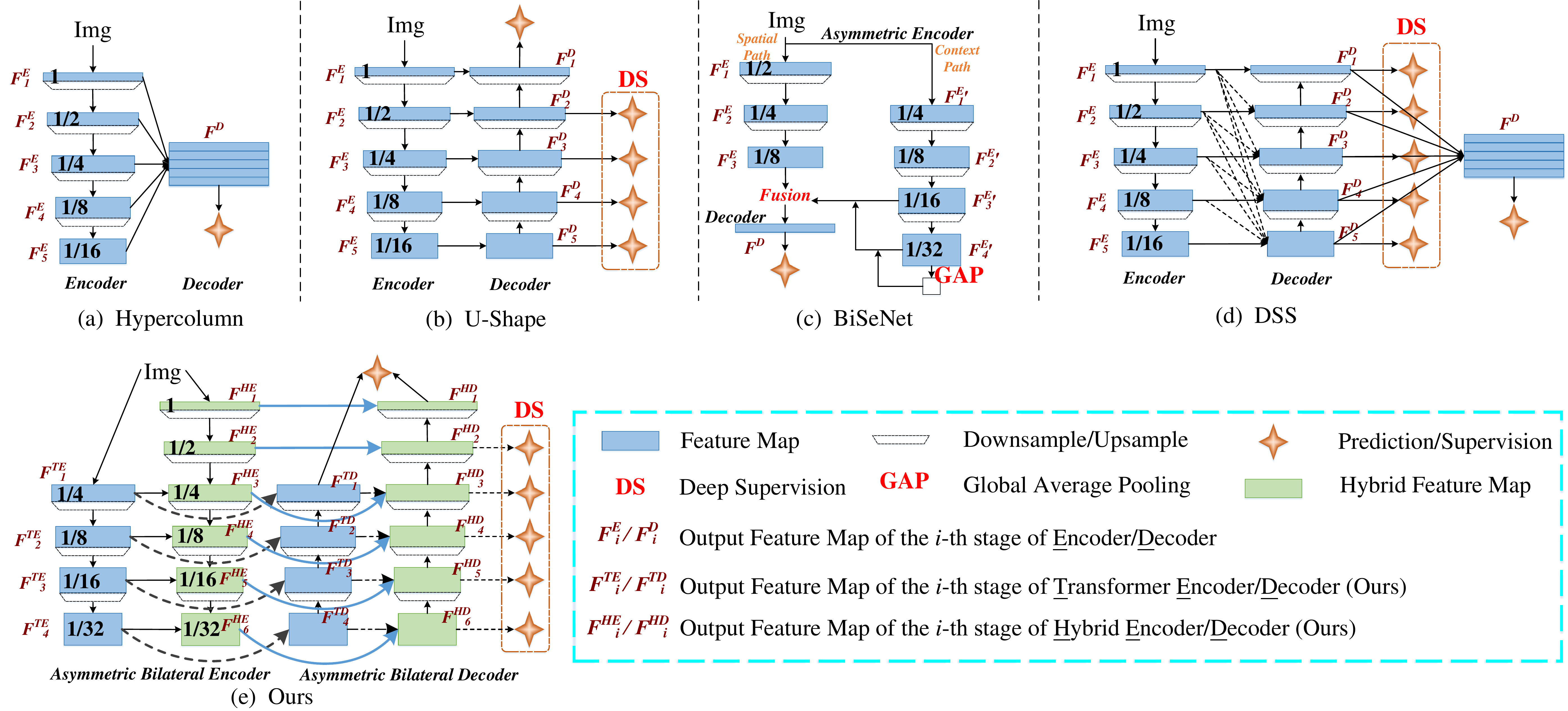}
    \vspace{-0.1in}
    \caption{Illustration of various encoder-decoder architectures. 
    (a) $\sim$ (e) indicate the architectures of
    Hypercolumn \cite{hariharan2015hypercolumns,li2016deep,wang2017stagewise,zeng2018learning,zhao2019pyramid,su2019selectivity,mao2021transformer},
    U-shape \cite{ronneberger2015u,wu2019cascaded,liu2022poolnet+,feng2019attentive,xu2019structured,zhao2019egnet,zhou2020interactive,pang2020multi,zhao2020suppress,chen2020global,liu2021visual}, BiSeNet \cite{yu2018bisenet},
    DSS \cite{xie2015holistically,hou2019deeply} and our ABiU-Net, respectively. 
    Note that the feature maps in (a) $\sim$ (d) (\ie, the blue rectangle) are not limited to transformer- or CNN-based feature maps.}
    \label{fig:arch}
    \vspace{-0.1in}
\end{figure*}

Based on the above observation, how to achieve effective local 
representations in accompany with transformer networks would be the key to further boost SOD.
To this end, we consider combining the merits of transformers 
and CNNs, which are adept at global relationship modeling and 
local representation learning, respectively.
In this paper, we propose a new encoder-decoder architecture, 
namely \underline{\textbf{A}}symmetric \underline{\textbf{Bi}}lateral
\underline{\textbf{U-Net}} (\textbf{ABiU-Net}).
Its encoder consists of two parts:
\underline{\textbf{T}}ransformer \underline{\textbf{Enc}}oder 
\underline{\textbf{Path}} (\textbf{TEncPath}) and
\underline{\textbf{H}}ybrid \underline{\textbf{Enc}}oder 
\underline{\textbf{Path}} (\textbf{HEncPath}).
TEncPath directly uses a transformer network for global
relationship modeling, in order to locate salient objects.
Since the transformer entirely relies on self-attention to extract 
global contextual features, TEncPath lacks local find-grained features 
that are essential for refining object details/boundaries.
Hence, we introduce HEncPath by stacking several convolution stages, to enhance the locality of the encoder.
To fuse the global and local features, the inputs of each stage of HEncPath are from the preceding convolution
stage as well as the corresponding TEncPath stage, respectively.
In this way, HEncPath introduces locality into feature representations
with the guide of global contexts from TEncPath.
Therefore, HEncPath is a hybrid encoding of global long-range 
dependencies and local representations.
Note that HEncPath is lightweight with a small number of channels and a fast downsampling strategy.

In the phase of decoding, we design an asymmetric 
bilateral decoder containing two simple paths, namely 
\underline{\textbf{T}}ransformer \underline{\textbf{Dec}}oder 
\underline{\textbf{Path}} (\textbf{TDecPath}) 
and \underline{\textbf{H}}ybrid \underline{\textbf{Dec}}oder 
\underline{\textbf{Path}} (\textbf{HDecPath}), 
which are utilized to decode feature representations from TEncPath 
and HEncPath, respectively.
TDecPath decodes coarse salient object locations, while HDecPath 
is expected to further refine object details/boundaries.
We adopt the standard channel attention mechanism to enhance 
the feature representations from TEncPath and HEncPath. 
The output of TDecPath at each stage is fed into the corresponding 
stage of HDecPath so that these two decoder paths can communicate 
and learn complementary information, \ie, coarse locations and 
find-grained details, respectively.

Generally, ABiU-Net is an extension of the U-shaped architecture, and
it mainly aims to explore a balanced combination of transformers and CNNs to
address the challenge of accurately segmenting salient object in natural scenes.
Although transformer-based U-Net models have been explored, they usually
replace the encoder/decoder of classic U-Net with transformers or the simple
serial combination of CNNs and transformers \cite{chen2021transunet,cao2021swin}.
In contrast, this paper introduces a bilateral encoder-decoder structure and enables the
communication between the two paths, which is ignored by previous studies.
Besides, the existing multi-column architectures \cite{yu2018bisenet,lin2022ds} are usually
built in a separate way, \ie, only the output features of the various paths are fused.
This late fusion manner can not fully utilize the complementary advantages of
different paths. Compared to their late fusion manner, the multi-level communication
of ABiU-Net has the following advantage: the global features learned from the transformer
path can guide the learning of detailed features of the CNN path, and in turn, the detailed
features can promote the optimization of global features. In brief, we can
conclude our contribution and novelty as: we propose a new network architecture, ABiU-Net, to make
complementary use of global contextual features (for locating salient objects) and
local detailed representations (for refining object details) by exploring the deep
cooperation between transformers and CNNs. Extensive experiments demonstrate 
the remarkable superiority of ABiU-Net. Considering that ABiU-Net is an elegant
architecture without carefully designed modules or engineering skills, ABiU-Net
provides a new perspective for SOD in the transformer era.


\section{Related Work}
\subsection{CNN-based Salient Object Detection}
In the last decade, the accuracy of SOD has been remarkably boosted due to the
multi-level representation capability of CNNs
\cite{zhang2017amulet,zhang2017learning,luo2017non,wang2017stagewise,liu2018picanet,li2018contour,chen2018reverse,liu2020lightweight,liu2021samnet,qiu2019revisiting,qiu2020simple,wu2019cascaded,zhao2019egnet,wang2020deep,liu2022poolnet+}.
It is widely accepted that the high-level semantic information
extracted by the top CNN layers is beneficial to locating the
coarse positions of salient objects, while the low-level information
extracted by the bottom layers can refine the object details.
Hence, both the high-level and low-level information are important
for accurate SOD
\cite{liu2016dhsnet,li2016deep,luo2017non,zhang2017learning,zhang2017amulet,wang2017stagewise,liu2018picanet,wang2018detect,hou2019deeply,zhang2022dhnet}.
Most existing CNN-based SOD methods use pre-trained image classification 
models, \eg, ResNet \cite{he2016deep},
as encoders and focus on designing effective decoders by aggregating
multi-level features
\cite{hou2019deeply,zhao2019pyramid,qiu2019revisiting,zhang2018progressive,feng2019attentive,liu2022poolnet+,zhang2022dhnet}.
For example, Wu \etal \cite{wu2022edn} introduced an extremely-downsampled network to show the importance of high-level features for SOD. 
Li \etal \cite{li2021dense} built dense attention upon multi-level features simultaneously for feature selection in SOD.
Mei \etal \cite{mei2021exploring} aimed at capturing dense multi-scale
contexts to enhance the feature discriminability.
Zhang \etal \cite{zhang2022tcrnet} proposed a novel trifurcated cascaded
refinement network to explore multi-level feature fusion and global
information representation for SOD. 
In addition, Liu \etal \cite{liu2020lightweight,liu2021samnet} 
also designed new encoders for lightweight SOD.

However, CNNs are limited in global dependency modeling which is 
essential for locating salient objects, as noted in \secref{sec:intro}.
To alleviate this issue, some modules are introduced
to learn the long-distance dependency information. 
For example, Hu \etal \cite{hu2020sac} designed a spatial attenuation 
context module to maximize the integration of local and global image
context within, around, and beyond the salient objects.
Gu \etal \cite{gu2020pyramid} introduced a pyramid constrained
self-attention operation to capture objects at various scales.
Qiu \etal \cite{qiu2022a2sppnet} designed attentive atrous
spatial pyramid pooling (A2SPP) by adding a new cubic 
information-embedding attention module at each branch of atrous
spatial pyramid pooling (ASPP) \cite{chen2017deeplab} 
to encode multi-level dependency information for SOD. 
A comprehensive review of CNN-based SOD is beyond the scope of 
this paper, and we recommend referring to \cite{wang2021salient}
for a more thorough analysis.

Due to the local receptive fields of convolution operations, 
the ability of the above CNN-based models in learning global semantic 
information is limited, which is the main bottleneck 
for improving CNN-based SOD.
To solve this problem, we note that recent vision transformers 
\cite{vaswani2017attention,dosovitskiy2021image}
are adept at global relationship modeling. 
Thus, we attempt to explore vision transformers for boosting SOD 
and delve into the effective cooperation of transformers and CNNs
in learning both global contexts and local features.

\subsection{Transformer-based Salient Object Detection}
The transformer is first proposed in NLP for machine translation \cite{vaswani2017attention}.
The transformer network alternately stacks multi-head self-attention modules, aiming at estimating the global dependencies between every two patches, and a multilayer perceptron, aiming at feature enhancement.
Recently, researchers have brought the transformer into computer vision and achieved remarkable achievements.
Specifically, Dosovitskiy \etal \cite{dosovitskiy2021image} made the first attempt to apply the transformer to image classification, attaining competitive performance on the ImageNet dataset \cite{russakovsky2015imagenet}.
They split an image into a sequence of flattened patches that are fed into transformers.
Following \cite{dosovitskiy2021image}, lots of studies have emerged in a short period, and much better performance than \sArt CNNs has been achieved \cite{yuan2021tokens,wang2021pyramid,liu2021swin,wu2022p2t,han2021transformer}.
For example, Cui \cite{cui2022dynamic} first proposed a dynamic
aggregation module to adaptively select the frames for feature enhancement,
which improved the inference speed of feature aggregation-based video
object detectors. Then, to get better performance, Cui \cite{cui2023feature}
focused on transformer-based video object detection and introduced a query
aggregation module to improve the quality of queries by aggregating queries according to the features of input frames.

For SOD, Mao \etal \cite{mao2021transformer} adopted Swin Transformer \cite{liu2021swin} as the encoder and designed a simple CNN-based decoder to predict saliency maps recently.
However, compared to CNN-based salient object detection models, they only replaced the CNN-based encoder with the transformer-based encoder.
To further explore the superiority of vision transformers in SOD, Liu \etal \cite{liu2021visual} introduced a novel pure transformer model for SOD, bringing a great success.
They designed a new token upsampling strategy and fused multi-level patch tokens to make transformer more suitable for SOD.
Moreover, Mao \etal \cite{mao2021generative} claimed that, compared to the CNN-based models, the superior performance of transformer-based models came from the effective global context modeling abilities.
Although existing transformer-based works \cite{hu2020sac,mao2021transformer,liu2021visual} solve the lack of global contexts in CNN-based SOD, they ignore the local representations that are essential for refining salient object details. 
In this paper, we work on how to effectively learn both global contexts and local features by exploring the cooperation of transformers and CNNs, so salient objects can be precisely located and segmented.

The proposed ABiU-Net uses the PVT \cite{wang2021pyramid}, for global relationship modeling.
In detail, PVT aims to introduce the pyramid structure into the transformer framework, which can generate
multi-scale feature maps for dense prediction.
PVT has four stages, each of which is comprised of a patch embedding layer and a $L_i$-layer transformer encoder.
To obtain a pyramid structure, PVT uses a progressive shrinking strategy to control the scale of feature maps by
patch embedding layers.  In this way, PVT can generate feature maps of four different scales, \ie, 4-. 8-, 16-,
and 32-stride with respect to the input image. PVT has a series of versions with different scales, namely PVT-Tiny,
PVT-Small, PVT-Medium, and PVT-Large, whose numbers of parameters are similar to ResNet18, ResNet50, ResNet101, and ResNet152 \cite{he2016deep}, respectively. 
By default, the TEncPath of ABiU-Net uses PVT-Small.

\subsection{Encoder-decoder Architectures}
This paper mainly designs a new encoder-decoder architecture, \ie, Asymmetric Bilateral U-Net (ABiU-Net), for SOD.
To illustrate the novelty of ABiU-Net, we summarize various widely-used encoder-decoder architectures (not only for SOD) in \figref{fig:arch}.
%
%
Hypercolumn \cite{hariharan2015hypercolumns} simply aggregates features from different levels of the encoder for final predictions, as shown in \figref{fig:arch}(a).
The aggregated hyper-features are so-called Hypercolumns.
The typical applications of the Hypercolumn encoder-decoder architecture in SOD include \cite{li2016deep,wang2017stagewise,zeng2018learning,zhao2019pyramid,su2019selectivity,mao2021transformer}.
As shown in \figref{fig:arch}(b), the U-shaped encoder-decoder architecture is the most widely-used. 
The main idea of U-shaped encoder-decoder is to supplement a contracting encoder with a symmetric decoder, where the pooling operations in the encoder are replaced with upsampling operations in the decoder.
Deep supervision \cite{lee2015deeply} is also imposed to ease the training process.
Most SOD methods are based on or improve upon the U-shaped encoder-decoder architecture \cite{zhang2017amulet,zhang2017learning,luo2017non,wang2017stagewise,wang2018detect,liu2018picanet,li2018contour,chen2018reverse,qiu2019revisiting,wu2019cascaded,feng2019attentive,xu2019structured,zhao2019egnet,zhou2020interactive,pang2020multi,zhao2020suppress,chen2020global,qiu2020simple,tu2020edge,liu2020lightweight,liu2021samnet,liu2021dna,liu2021visual,liu2022poolnet+}.
As shown in \figref{fig:arch}(c), BiSeNet \cite{yu2018bisenet}, designed for semantic segmentation, has two encoder paths: (1) the spatial path that stacks only three convolution layers to obtain the $1/8$ feature map to retain affluent spatial details, and (2) the context path connects a global average pooling layer at the top. 
These two paths are fused for final prediction.
Note that our ``bilateral'' is inspired by BiSeNet \cite{yu2018bisenet}. 
Hence, we discuss the difference between the proposed model and BiSeNet here.
On one hand, although both BiSeNet and ABiU-Net have a two-path encoder, ABiU-Net enables interaction between the two paths, while BiSeNet does not. 
On the other hand, ABiU-Net supplements an asymmetric bilateral decoder for better fusing the information from the two-path encoder, while BiSeNet directly utilizes the features of the encoder for final prediction.
Another widely-used encoder-decoder architecture is Holistically-nested Edge Detector (HED) \cite{xie2015holistically}.
Hou \etal \cite{hou2019deeply} designed a new HED-based saliency detection model, namely DSS, by introducing short connections to the skip-layer structure within the HED, which is shown in \figref{fig:arch}(d).
From \figref{fig:arch}, we can clearly see the difference between the proposed ABiU-Net and other architectures.
Therefore, we can conclude that ABiU-Net is a new encoder-decoder architecture in the transformer era.

\begin{figure*}[!tb]
    \centering
    \includegraphics[width=\linewidth]{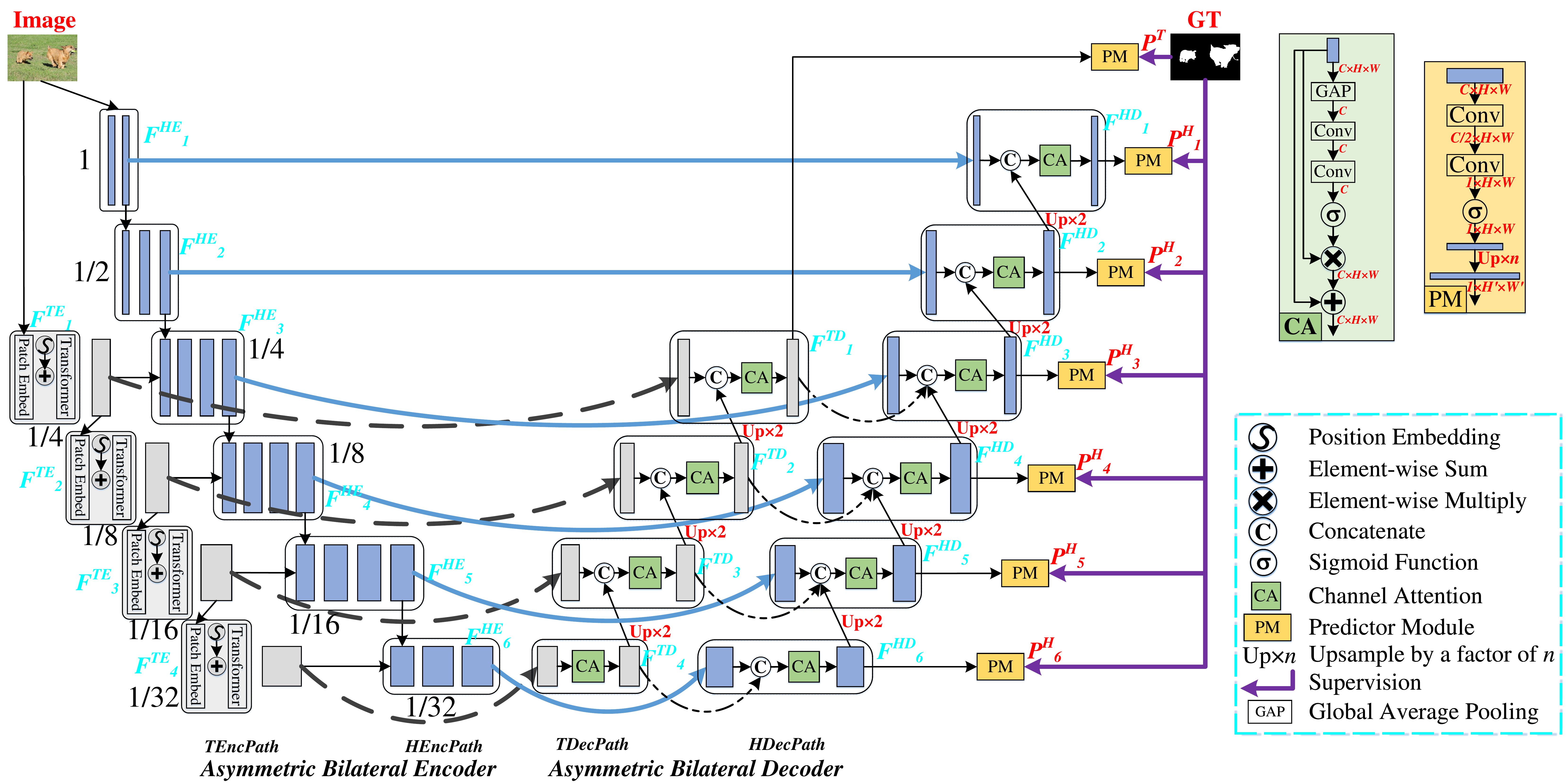}
    \vspace{-0.1in}
    \caption{Framework of the proposed Asymmetric Bilateral U-Net (ABiU-Net).}
    \label{fig:frame}
\end{figure*}

\section{Methodology} \label{ssec:method}
This section presents the proposed ABiU-Net for accurate SOD.
We first describe the overall framework of ABiU-Net 
in \secref{sec:framework}.
Then, we introduce the asymmetric bilateral encoder and the asymmetric bilateral decoder in \secref{sec:encoder} and \secref{sec:decoder}, respectively.

\subsection{Overall Framework}\label{sec:framework}
Since previous CNN-based SOD methods 
\cite{wang2015deep,lee2016deep,wang2016saliency,li2016deep,zhang2017amulet,zhang2017learning,luo2017non,wang2017stagewise,liu2018picanet,li2018contour,chen2018reverse,liu2020lightweight,liu2021samnet,zhang2021engaging}
lack powerful global context modeling 
and existing transformer-based SOD methods 
\cite{mao2021transformer,liu2021visual}
lack effective local representations, 
in this paper, we aim at learning both global contexts and 
local features for locating salient objects and refining
object details, respectively.
For this goal, we improve the traditional U-Net \cite{ronneberger2015u}
to the new ABiU-Net by exploring the cooperation of transformers and CNNs, so ABiU-Net can inherit their merits for global context modeling and local feature learning, respectively.
ABiU-Net consists of an asymmetric bilateral encoder and an asymmetric bilateral decoder.

As shown in \figref{fig:frame}, the asymmetric bilateral encoder of ABiU-Net contains two parts: TEncPath in grey and HEncPath in blue.
TEncPath is one well-known transformer network, \eg, 
PVT \cite{wang2021pyramid}, with four stages. 
The main issue is that the transformer network only focuses 
on learning long-range dependencies with the sacrifice of 
local information.
This is remedied by another encoder path, \ie, HEncPath.
HEncPath stacks six lightweight convolution stages. 
In particular, the inputs of the $3^{\rm rd}$ $\sim$ $6^{\rm th}$ 
stages are not only from the preceding stage but also from the 
corresponding transformer stage that has the same output stride.
Hence, HEncPath can achieve the hybrid encoding of global 
long-range dependencies and local representations.

The asymmetric bilateral decoder also consists of two paths, 
namely TDecPath and HDecPath, decoding feature representations 
from TEncPath and HEncPath, respectively.
TDecPath has four stages, which can be viewed as a simple
top-down generation path to regress the coarse locations of
salient objects.
HDecPath is a hybrid decoding process with six stages 
in the top-down view.
The input of HDecPath at each stage is not only from HEncPath
but also from TDecPath.
In this way, two decoder paths can communicate and learn 
complementary information, \ie, the coarse locations of salient 
objects decoded in TDecPath would guide the learning of 
HDecPath to further refine object details.
We will introduce the decoder in \secref{sec:decoder}.

The output feature map of HDecPath at the last stage is used
for the final saliency map prediction.
We also impose deep supervision \cite{lee2015deeply} on the other 
five stages of HDecPath and the last stage of TDecPath.
To achieve this, we design a simple \textbf{Prediction Module (PM)}, 
which converts a feature map to a saliency map.
PM first adopts two successive $3\times 3$ convolution layers 
with batch normalization and nonlinearization to convert the input
feature map to a single-channel map.
Then, the \textit{sigmoid} activation function is followed to
predict the saliency probability map, whose values range from 0 to 1.
The proposed ABiU-Net is trained end-to-end using the standard
\textit{binary cross-entropy loss} (BCE).
Suppose the saliency maps of HDecPath are denoted by 
$\mathbf{P}^H_i$ ($i\in {1,2,3,4,5,6}$) from bottom to top,
and the saliency map of TDecPath is denoted by $\mathbf{P}^T$.
The total training loss can be calculated as 
\begin{equation}
L = \mathcal{L}_{\rm BCE}(\mathbf{P}^H_1, \mathbf{G}) + \lambda \sum_{i=2}^{6} \mathcal{L}_{\rm BCE}(\mathbf{P}^H_i, \mathbf{G}) + \lambda \mathcal{L}_{\rm BCE}(\mathbf{P}^T, \mathbf{G}),
\end{equation}
where $\mathbf{G}$ represents the ground-truth saliency map.
$\lambda$ is a weighting scalar for loss balance. 
In this paper, we empirically set $\lambda$
to 0.4, as suggested by 
\cite{zhao2017pyramid,liu2021dna,liu2020lightweight,liu2021samnet,qiu2019revisiting,qiu2020simple}.
During testing, $\mathbf{P}^H_1$ is viewed as the final output 
saliency map.

\subsection{Asymmetric Bilateral Encoder}\label{sec:encoder}
First of all, we want to clarify that the asymmetry of our bilateral 
encoder mainly refers to:
i) two encoder paths are based on different networks 
(transformer and CNN);
ii) the numbers of stages of these two paths are different 
(four for TEncPath and six for HEncPath); and 
iii) the targets that they are responsible for are different 
(global context modeling and hybrid feature encoding).
Now, we describe how we designed it and the reasons behind it.

We use the popular transformer network, PVT \cite{wang2021pyramid},
as TEncPath.
All stages of PVT share a similar architecture, which consists of 
a patch embedding layer and several transformer blocks.
Specifically, given an input image, PVT splits it into small patches 
with the size of $4\times 4$, using a patch embedding layer.
Then, the flattened patches are added with a position embedding 
and fed into transformer blocks.
From the second stage, PVT utilizes a patch embedding layer
to shrink the feature map by a scale of 2 at the beginning 
of each stage, followed by the addition with a position embedding
and then some transformer blocks.
Suppose $\mathbf{F}^{\rm TE}_1$, $\mathbf{F}^{\rm TE}_2$,
$\mathbf{F}^{\rm TE}_3$, and $\mathbf{F}^{\rm TE}_4$ denote
the output feature maps of the four stages of PVT 
from bottom to top, and they have scales of
$1/4$, $1/8$, $1/16$, and $1/32$ with 64, 128, 320, and 512
channels, respectively.
Please refer to the original paper \cite{wang2021pyramid} 
for more details.

For HEncPath, we design a lightweight CNN-based sub-network to 
introduce the local sensitivity.
Specifically, we stack six convolution stages whose outputs are
$\mathbf{F}^{\rm HE}_1$, $\mathbf{F}^{\rm HE}_2$, 
$\mathbf{F}^{\rm HE}_3$, $\mathbf{F}^{\rm HE}_4$, 
$\mathbf{F}^{\rm HE}_5$, and $\mathbf{F}^{\rm HE}_6$ 
from bottom to top, respectively.
Except for the last stage, a max-pooling layer with a stride of 2 
is connected after each stage for feature downsampling,
leading to output scales of $1$, $1/2$, $1/4$, $1/8$, $1/16$, 
and $1/32$ for six stages from bottom to top, respectively.
The input image is fed into not only TEncPath but also HEncPath.
The output of the first stage of HEncPath is used as the input 
of its second stage.
From the third stage, the input of each stage is the concatenation 
of the feature map from the preceding stage and the feature map 
from the corresponding stage of TEncPath with the same resolution.
The concatenated feature map is first connected to a $1 \times 1$ 
convolution for integration.
Then, HEncPath processes it through $N_i$ convolution layers
with batch normalization and nonlinearization.
Through using $\mathbf{F}^{\rm TE}_i$ as the input of HEncPath,
it is easier for $\mathbf{F}^{\rm HE}_i$ to learn complementary
local fine-grained features with the guidance of 
$\mathbf{F}^{\rm TE}_i$.
Since TEncPath provides a high-level semantic abstraction for 
HEncPath, it is unnecessary to use a very deep or cumbersome 
CNN for HEncPath.
Note that HEncPath also takes the original image as input
so as to mine complementary local information from the image.

For a clear presentation, we can formulate these steps of HEncPath as
\begin{equation}
\begin{aligned}
\mathbf{F}^{\rm HE}_1 &= {\rm Conv}^{3\times 3}_{\times N_1}(\mathbf{I}), \\
\mathbf{F}^{\rm HE}_2 &= {\rm Conv}^{3\times 3}_{\times N_2}({\rm MaxPool}(\mathbf{F}^{\rm HE}_1)), \\
\mathbf{F}^{\rm HE}_i &= {\rm Conv}^{3\times 3}_{\times N_i}({\rm Conv}^{1\times 1}({\rm MaxPool}(\mathbf{F}^{\rm HE}_{i-1})\ \copyright\ \mathbf{F}^{\rm TE}_{i-2})), \\
    & \qquad\qquad\qquad\text{for}\ i \in \{3,4,5,6\},
\end{aligned}
\end{equation}
where $\mathbf{I}$ denotes the input color image.
${\rm Conv}^{1\times 1}(\cdot)$ is $1\times 1$ convolution.
${\rm Conv}^{3\times 3}_{\times N_i}(\cdot)$ represents
$N_i$ successive $3\times 3$ convolutions with 
batch normalization and nonlinearization omitted for simplicity,
where $N_i$ is the number of convolution layers at each stage 
of HEncPath.
${\rm MaxPool}(\cdot)$ is a max pooling layer with a stride of 2.
``$\copyright$'' represents the concatenation operation along 
the channel dimension.
We set $N_i$ to 2 for $i\in \{1,2,3,4,5\}$, and we have $N_6=3$,
with 16, 64, 64, 128, 256, and 256 output channels, respectively,
to make HEncPath a lightweight sub-network.

With the asymmetric bilateral encoder, we obtain two sets of 
features with different characteristics.
$\mathbf{F}^{\rm TE}_i$ ($i\in {1,2,3,4}$), generated by 
TEncPath, is based on global long-range dependence modeling, 
thus containing rich contextual information.
On the other hand, $\mathbf{F}^{\rm HE}_i$ ($i\in {1,2,3,4,5,6}$)
aims at learning complementary information to global modeling,
guided by $\mathbf{F}^{\rm TE}_i$.
Hence, $\mathbf{F}^{\rm HE}_i$ contains rich local features.
The transformer features, $\mathbf{F}^{\rm TE}_i$, would be useful 
to locate salient objects with the global view on the image scenes, 
while the CNN features, $\mathbf{F}^{\rm HE}_i$, would be useful 
to refine object details with the local fine-grained 
representations.
Therefore, the combination of 
the global features $\mathbf{F}^{\rm TE}_i$ and
the local features $\mathbf{F}^{\rm HE}_i$ 
would lead to accurate SOD.

\subsection{Asymmetric Bilateral Decoder}\label{sec:decoder}
Corresponding to the encoder, we design an asymmetric bilateral 
decoder containing two paths, \ie, TDecPath and HDecPath.
TDecPath is utilized to decode feature representations
from TEncPath, which can be viewed as a simple top-down
generation path.
The top stage takes $\mathbf{F}^{\rm TE}_4$ as input,
and its output is denoted as $\mathbf{F}^{\rm TD}_4$.
The $i$-th ($i\in \{3,2,1\}$) stage of TDecPath has two inputs, 
\ie, $\mathbf{F}^{\rm TE}_i$ from TEncPath and 
$\mathbf{F}^{\rm TD}_{i+1}$ from the preceding stage of TDecPath,
generating the output $\mathbf{F}^{\rm TD}_i$.
The main operations of TDecPath are as below.
First of all, $\mathbf{F}^{\rm TE}_4$ $\sim$ $\mathbf{F}^{\rm TE}_1$
are separately fed into a $3\times 3$ convolution layer
with batch normalization and nonlinearization to reduce 
the number of channels to (128, 64, 32, 16), 
generating feature maps 
$\hat{\mathbf{F}}^{\rm TE}_i$ ($i \in \{4,3,2,1\}$), respectively.
This process can be formulated as
\begin{equation}
\hat{\mathbf{F}}^{\rm TE}_i = {\rm Conv}^{3\times 3}(\mathbf{F}^{\rm TE}_i),
\quad\text{for}\ i \in \{4,3,2,1\}.
\end{equation}
After that, $\hat{\mathbf{F}}^{\rm TE}_i$ ($i \in \{3,2,1\}$)
is concatenated with the output of the preceding stage in TDecPath.
Note that $\hat{\mathbf{F}}^{\rm TE}_4$ is the only input of the 
fourth stage of TDecPath, so there is no concatenation operation.
Then, we adopt the standard \textbf{Channel Attention (CA)}
mechanism to process the concatenated feature map 
for feature enhancement.
As shown in \figref{fig:frame}, the CA mechanism is a typical 
squeeze-excitation attention block \cite{hu2018squeeze}
whose description is omitted here.
After the CA block, we obtain $\mathbf{F}^{\rm TD}_4$,
$\mathbf{F}^{\rm TD}_3$, $\mathbf{F}^{\rm TD}_2$, 
and $\mathbf{F}^{\rm TD}_1$ from top to bottom.
Formally, this can be written as 
\begin{equation}
\begin{aligned}
\mathbf{F}^{\rm TD}_4 &= {\rm CA}(\hat{\mathbf{F}}^{\rm TE}_4),\\
\tilde{\mathbf{F}}^{\rm TD}_i &= {\rm Conv}^{3\times 3}(\mathbf{F}^{\rm TD}_{i}),
\quad\text{for}\ i \in \{4,3,2,1\},\\
\mathbf{F}^{\rm TD}_i &= {\rm CA}(\hat{\mathbf{F}}^{\rm TE}_i\ \copyright\ {\rm Upsample}(\tilde{\mathbf{F}}^{\rm TD}_{i+1})),
\quad\text{for}\ i \in \{3,2,1\},
\end{aligned}
\end{equation}
where the $3 \times 3$ convolution reduces the number
of feature channels of 
$\mathbf{F}^{\rm TD}_i$ ($i \in \{4,3,2,1\}$) to the half.
${\rm Upsample}(\cdot)$ is to upsample the feature map
by a factor of 2.
In this way, TDecPath can decode the coarse locations of 
salient objects using the global contexts in TEncPath.

HDecPath is expected to further refine salient object details, 
guided by the coarse locations decoded by TDecPath.
Hence, HDecPath takes two inputs: the features from TDecPath 
and HEncPath, providing the coarse locations of salient objects
and the features about object details, respectively.
HDecPath contains six decoder stages, whose outputs are
denoted as $\mathbf{F}^{\rm HD}_i$ ($i \in \{1,2,3,4,5,6\}$).
First, we connect the $3 \times 3$ convolution layers 
(with batch normalization and nonlinearization) 
to the side-outputs of HEncPath and TDecPath,
generating $\hat{\mathbf{F}}^{\rm HE}_i$ and 
$\hat{\mathbf{F}}^{\rm TD}_i$, respectively.
This can be formulated as 
\begin{equation}
\begin{aligned}
\hat{\mathbf{F}}^{\rm HE}_i &= {\rm Conv}^{3\times 3}(\mathbf{F}^{\rm HE}_i),
\quad\text{for}\ i \in \{6,5,4,3,2,1\}, \\
\hat{\mathbf{F}}^{\rm TD}_i &= {\rm Conv}^{3\times 3}(\mathbf{F}^{\rm TD}_i),
\quad\text{for}\ i \in \{4,3,2,1\},
\end{aligned}
\end{equation}
in which $\hat{\mathbf{F}}^{\rm HE}_i$ ($i \in \{6,5,4,3,2,1\}$)
has (256, 128, 64, 32, 32, 8) channels, respectively.
$\hat{\mathbf{F}}^{\rm TD}_i$ ($i \in \{4,3,2,1\}$) has 
the same number of channels as $\hat{\mathbf{F}}^{\rm TE}_i$.
Note that $\hat{\mathbf{F}}^{\rm HE}_{i+2}$ and 
$\hat{\mathbf{F}}^{\rm TD}_i$ ($i \in \{4,3,2,1\}$)
have the same scale.

Then, we feed $\hat{\mathbf{F}}^{\rm HE}_6$ and 
$\hat{\mathbf{F}}^{\rm TD}_4$ into a CA block
for feature fusion, followed by a $3 \times 3$ convolution
to produce the refined feature $\mathbf{F}^{\rm HD}_6$.
Different from the sixth stage, 
the $i$-th stage ($i \in \{5,4,3\}$) has three inputs,
\ie, $\hat{\mathbf{F}}^{\rm HE}_i$, $\hat{\mathbf{F}}^{\rm TD}_{i-2}$,
and $\mathbf{F}^{\rm HD}_{i+1}$, where $\mathbf{F}^{\rm HD}_{i+1}$
should be upsampled by a factor of 2 first.
These three inputs are concatenated, whose result is 
fed into a CA block.
After that, a $3 \times 3$ convolution is connected for feature fusion,
generating the output $\mathbf{F}^{\rm HD}_i$ ($i \in \{5,4,3\}$).
However, for the first and second stages of HDecPath, 
the operations are the same as TDecPath because there are 
no side-outputs from TDecPath with the same scales.
The inputs of these two stages are $\hat{\mathbf{F}}^{\rm HE}_i$ 
and $\mathbf{F}^{\rm HD}_{i+1}$ ($i \in \{2,1\}$),
and the outputs are $\mathbf{F}^{\rm HD}_i$ ($i \in \{2,1\}$).
We formulate the computation process of HDecPath as
\begin{equation}\label{equ:HDecPath}
\begin{aligned}
\mathbf{F}^{\rm HD}_6 &= {\rm Conv}^{3\times 3}({\rm CA}(\hat{\mathbf{F}}^{\rm TD}_4 \ \copyright\ \hat{\mathbf{F}}^{\rm HE}_6)), \\
\tilde{\mathbf{F}}^{\rm HD}_i &= {\rm Upsample}(\mathbf{F}^{\rm HD}_i),
\quad\text{for}\ i \in \{6,5,4,3,2\}, \\
\mathbf{F}^{\rm HD}_i &= {\rm Conv}^{3\times 3}({\rm CA}(\hat{\mathbf{F}}^{\rm TD}_{i-2} \copyright \tilde{\mathbf{F}}^{\rm HD}_{i+1} \copyright \hat{\mathbf{F}}^{\rm HE}_i)), 
\text{for}\ i \in \{5,4,3\}, \\
\mathbf{F}^{\rm HD}_i &= {\rm Conv}^{3\times 3}({\rm CA}(\tilde{\mathbf{F}}^{\rm HD}_{i+1}\ \copyright\ \hat{\mathbf{F}}^{\rm HE}_i)), 
\quad\text{for}\ i \in \{2,1\}. \\
\end{aligned}
\end{equation}
With the $3\times 3$ convolutions in \equref{equ:HDecPath},
HDecPath produces the final decoded feature maps 
$\mathbf{F}^{\rm HD}_i$ ($i \in \{6,5,4,3,2,1\}$)
with (128, 64, 32, 32, 8, 8) channels, respectively.

The term ``bilateral'' means that a network has two paths, like the well-known BiSeNet \cite{yu2018bisenet} which consists of two isolated CNN paths (\ie, the simple spatial path and the context path).
In this paper, the encoder of ABiU-Net consists of two connected paths: Transformer Encoder Path (\textbf{TEncPath}) and Hybrid Encoder Path (\textbf{HEncPath}), which focus on modeling global relationships and learning complementary local fine-grained representations, respectively.
Hence, we simply name the proposed encoder path as \textit{asymmetric bilateral encoder}.
Different from BiSeNet \cite{yu2018bisenet} which directly adopts the output features from the encoder for the final prediction, our ABiU-Net also contains an \textit{asymmetric bilateral decoder} for decoding object locations and object details from the encoding global and local features, respectively.
As a result, ABiU-Net contributes a new perspective on the design of transformer networks for SOD, while BiSeNet \cite{yu2018bisenet} is a pure CNN model.

\begin{table*}[!t]
\centering
\setlength{\tabcolsep}{3.0mm}
\caption{Comparison between the proposed ABiU-Net and state-of-the-art methods in terms of  $F_\beta$ ($\uparrow$), MAE ($\downarrow$), $F_\beta^w$ ($\uparrow$), and $S_m$ ($\uparrow$) on six datasets. The best result in each column is highlighted in \textbf{bold}.}
\label{tab:eval_f_mae}
\resizebox{\linewidth}{!}{
\begin{tabular}{c|l||c|c|c|c|c|c|c|c|c|c|c|c} \hline
    \rowcolor{mycolor} 
    & & \multicolumn{2}{c|}{SOD} & \multicolumn{2}{c|}{HKU-IS}
    & \multicolumn{2}{c|}{ECSSD} & \multicolumn{2}{c|}{DUT-OMRON}
    & \multicolumn{2}{c|}{THUR15K} & \multicolumn{2}{c}{DUTS-test}
    \\ \cline{3-14} \rowcolor{mycolor}
    \multirow{-2}{*}{\#} & \multirow{-2}{*}{Methods}
    & $F_\beta$ & MAE & $F_\beta$ & MAE & $F_\beta$ & MAE
    & $F_\beta$ & MAE & $F_\beta$ & MAE & $F_\beta$ & MAE
    \\ \hline
    1 & UCF$_{\textit{2017}}$ \cite{zhang2017learning}
    & .805 & .148 & .888 & .062 & .901 & .071
    & .730 & .120 & .758 & .112 & .772 & .112 
    \\ \rowcolor{mycolor}
    2 & SRM$_{\textit{2017}}$ \cite{wang2017stagewise}
    & .840 & .126 & .906 & .046 & .914 & .056
    & .769 & .069 & .778 & .077 & .826 & .059 
    \\
    3 & PiCA$_{\textit{2018}}$ \cite{liu2018picanet}
    & .836 & .102 & .916 & .042 & .923 & .049
    & .766 & .068 & .783 & .083 & .837 & .054
    \\ \rowcolor{mycolor}
    4 & BRN$_{\textit{2018}}$ \cite{wang2018detect}
    & .843 & .103 & .910 & .036 & .919 & .043
    & .774 & .062 & .769 & .076 & .827 & .050 
    \\
    5 & C2S$_{\textit{2018}}$ \cite{li2018contour}
    & .819 & .122 & .898 & .046 & .907 & .057
    & .759 & .072 & .775 & .083 & .811 & .062 
    \\ \rowcolor{mycolor}
    6 & RAS$_{\textit{2018}}$ \cite{chen2018reverse}
    & .847 & .123 & .913 & .045 & .916 & .058
    & .785 & .063 & .772 & .075 & .831 & .059 
    \\
    7 & DSS$_{\textit{2019}}$ \cite{hou2019deeply}
    & .842 & .122 & .913 & .041 & .915 & .056
    & .774 & .066 & .770 & .074 & .827 & .056
    \\ \rowcolor{mycolor}
    8 & PAGE-Net$_{\textit{2019}}$ \cite{wang2019salient}
    & .837 & .110 & .918 & .037 & .927 & .046
    & .791 & .062 & .766 & .080 & .838 & .052
    \\
    9 & AFNet$_{\textit{2019}}$ \cite{feng2019attentive}
    & .848 & .108 & .921 & .036 & .930 & .045
    & .784 & .057 & .791 & .072 & .857 & .046
    \\ \rowcolor{mycolor}
    10 & DUCRF$_{\textit{2019}}$ \cite{xu2019structured}
    & .836 & .121 & .920 & .040 & .924 & .052
    & .802 & .057 & .762 & .080 & .833 & .059
    \\
    11 & HRSOD$_{\textit{2019}}$ \cite{zeng2019towards}
    & .819 & .138 & .912 & .042 & .916 & .058
    & .752 & .066 & .784 & .068 & .836 & .051
    \\ \rowcolor{mycolor}
    12 & CPD$_{\textit{2019}}$ \cite{wu2019cascaded}
    & .848 & .113 & .924 & .033 & .930 & .044
    & .794 & .057 & .795 & .068 & .861 & .043
    \\
    13 & BASNet$_{\textit{2019}}$ \cite{qin2019basnet}
    & .849 & .112 & .928 & .032 & .938 & .040
    & .805 & .056 & .783 & .073 & .859 & .048
    \\ \rowcolor{mycolor}
    14 & EGNet$_{\textit{2019}}$ \cite{zhao2019egnet}
    & .859 & .110 & .928 & .034 & .938 & .044
    & .794 & .056 & .800 & .070 & .870 & .044
    \\
    15 & F$^3$Net$_{\textit{2020}}$ \cite{wei2020f3net}
    & .857 & .104 & .929 & .032 & .940 & .040
    & .802 & .059 & .795 & .069 & .859 & .044
    \\ \rowcolor{mycolor}
    16 & ITSD$_{\textit{2020}}$ \cite{zhou2020interactive}
    & .867 & .098 & .926 & .035 & .939 & .040
    & .802 & .063 & .806 & .068 & .875 & .042
    \\
    17 & MINet$_{\textit{2020}}$ \cite{pang2020multi}
    & .842 & .099 & .929 & .032 & .937 & .040
    & .780 & .057 & .808 & .066 & .870 & .040
    \\ \rowcolor{mycolor}
    18 & LDF$_{\textit{2020}}$ \cite{wei2020label}
    & .863 & .101 & .935 & .028 & .939 & .041
    & .803 & .057 & .815 & .064 & .886 & .039
    \\
    19 & GCPANet$_{\textit{2020}}$ \cite{chen2020global}
    & .842 & .100 & .935 & .032 & .942 & .037
    & .796 & .057 & .803 & .066 & .872 & .038
    \\ \rowcolor{mycolor}
    20 & GateNet$_{\textit{2020}}$ \cite{zhao2020suppress}
    & .851 & .108 & .927 & .036 & .933 & .045
    & .784 & .061 & .808 & .068 & .866 & .045
    \\
    21 & VST$_{\textit{2021}}$ \cite{liu2021visual}
    & .873 & \textbf{.083} & .942 & .030 & .951 & .034
    & .822 & .058 & .804 & .075 & .890 & .038
    \\ \rowcolor{mycolor}
    22 & DCENet$_{\textit{2021}}$ \cite{mei2021exploring}
    & .864 & .089 & .933 & .030 & .947 & .035
    & .806 & .056 & .817 & .065 & .882 & .038
    \\
    23 & PoolNet+$_{\textit{2022}}$ \cite{liu2022poolnet+}
    & .863 & .111 & .925 & .037 & .939 & .045
    & .791 & .060 & .800 & .068 & .866 & .043
    \\ \rowcolor{mycolor}
    24 & DNA$_{\textit{2022}}$ \cite{liu2021dna}
    & .851 & .113 & .928 & .036 & .940 & .043
    & .803 & .063 & .801 & .073 & .874 & .047
    \\
    25 & ICON$_{\textit{2022}}$ \cite{zhuge2022salient}
    & \textbf{.879} & .084 & .938 & .029 & .950 & .032
    & .821 & .057 & .805 & .076 & .886 & .040
    \\ \rowcolor{mycolor}
    26 & RCSBNet$_{\textit{2022}}$ \cite{ke2022recursive}
    & .869 & .085 & .938 & .027 & .944 & .034
    & .797 & .049 & .801 & .064 & .885 & .035
    \\
    27 & \textbf{ABiU-Net (ours)}
    & \textbf{.879} & .089 & \textbf{.951} & \textbf{.021} 
    & \textbf{.959} & \textbf{.026} & \textbf{.843} & \textbf{.043}
    & \textbf{.820} & \textbf{.059} & \textbf{.906} & \textbf{.029}
    \\ \hline\hline
    %
    \rowcolor{mycolor}
    & & \multicolumn{2}{c|}{SOD} & \multicolumn{2}{c|}{HKU-IS}
    & \multicolumn{2}{c|}{ECSSD} & \multicolumn{2}{c|}{DUT-OMRON}
    & \multicolumn{2}{c|}{THUR15K} & \multicolumn{2}{c}{DUTS-test}
    \\ \cline{3-14} \rowcolor{mycolor}
    \multirow{-2}{*}{\#} & \multirow{-2}{*}{Methods}
    & $F_\beta^w$ & $S_m$ & $F_\beta^w$ & $S_m$
    & $F_\beta^w$ & $S_m$ & $F_\beta^w$ & $S_m$
    & $F_\beta^w$ & $S_m$ & $F_\beta^w$ & $S_m$
    \\ \hline
    1 & UCF$_{\textit{2017}}$ \cite{zhang2017learning}
    & .673 & .763 & .779 & .875 & .805 & .884
    & .574 & .760 & .613 & .785 & .595 & .782
    \\ \rowcolor{mycolor}
    2 & SRM$_{\textit{2017}}$ \cite{wang2017stagewise}
    & .670 & .739 & .835 & .887 & .849 & .894
    & .658 & .798 & .684 & .818 & .721 & .836 
    \\
    3 & PiCA$_{\textit{2018}}$ \cite{liu2018picanet}
    & .721 & .787 & .847 & .905 & .862 & .914
    & .691 & .826 & .688 & .823 & .745 & .860 
    \\ \rowcolor{mycolor}
    4 & BRN$_{\textit{2018}}$ \cite{wang2018detect}
    & .670 & .768 & .835 & .895 & .849 & .902
    & .658 & .806 & .684 & .813 & .721 & .842
    \\
    5 & C2S$_{\textit{2018}}$ \cite{li2018contour}
    & .700 & .757 & .835 & .889 & .849 & .896
    & .663 & .799 & .685 & .812 & .717 & .832
    \\ \rowcolor{mycolor}
    6 & RAS$_{\textit{2018}}$ \cite{chen2018reverse}
    & .718 & .761 & .850 & .889 & .855 & .894
    & .695 & .812 & .691 & .813 & .739 & .839 
    \\ 
    7 & DSS$_{\textit{2019}}$ \cite{hou2019deeply}
    & .711 & .747 & .862 & .881 & .864 & .884
    & .688 & .790 & .702 & .805 & .752 & .826
    \\ \rowcolor{mycolor}
    8 & PAGE-Net$_{\textit{2019}}$ \cite{wang2019salient}
    & .721 & .769 & .865 & .903 & .879 & .912
    & .722 & .825 & .698 & .815 & .768 & .854
    \\
    9 & AFNet$_{\textit{2019}}$ \cite{feng2019attentive}
    & .726 & .773 & .869 & .905 & .880 & .913
    & .717 & .826 & .719 & .829 & .784 & .867
    \\ \rowcolor{mycolor}
    10 & DUCRF$_{\textit{2019}}$ \cite{xu2019structured}
    & .697 & .760 & .855 & .903 & .858 & .907
    & .706 & .821 & .663 & .801 & .723 & .836
    \\
    11 & HRSOD$_{\textit{2019}}$ \cite{zeng2019towards}
    & .622 & .702 & .851 & .882 & .853 & .883
    & .645 & .772 & .713 & .820 & .746 & .830
    \\ \rowcolor{mycolor}
    12 & CPD$_{\textit{2019}}$ \cite{wu2019cascaded}
    & .718 & .765 & .879 & .904 & .888 & .910
    & .715 & .818 & .730 & .831 & .799 & .866
    \\
    13 & BASNet$_{\textit{2019}}$ \cite{qin2019basnet}
    & .728 & .766 & .889 & .909 & .898 & .916
    & .741 & .836 & .721 & .823 & .802 & .865
    \\ \rowcolor{mycolor}   
    14 & EGNet$_{\textit{2019}}$ \cite{zhao2019egnet}
    & .736 & .782 & .876 & .912 & .886 & .919
    & .727 & .836 & .727 & .836 & .796 & .878
    \\
    15 & F$^3$Net$_{\textit{2020}}$ \cite{wei2020f3net}
    & .742 & .785 & .883 & .915 & .900 & .924
    & .710 & .823 & .726 & .835 & .790 & .872
    \\ \rowcolor{mycolor}
    16 & ITSD$_{\textit{2020}}$ \cite{zhou2020interactive}
    & .764 & .791 & .881 & .906 & .897 & .914
    & .734 & .829 & .739 & .836 & .813 & .887
    \\
    17 & MINet$_{\textit{2020}}$ \cite{pang2020multi}
    & .740 & .781 & .889 & .912 & .899 & .919
    & .719 & .822 & .742 & .838 & .812 & .874
    \\ \rowcolor{mycolor}
    18 & LDF$_{\textit{2020}}$ \cite{wei2020label}
    & .754 & .787 & .891 & .918 & .890 & .915
    & .715 & .826 & .727 & .835 & .807 & .879
    \\
    19 & GCPANet$_{\textit{2020}}$ \cite{chen2020global}
    & .731 & .776 & .889 & .918 & .900 & .922
    & .734 & .830 & .732 & .839 & .817 & .884
    \\ \rowcolor{mycolor}
    20 & GateNet$_{\textit{2020}}$ \cite{zhao2020suppress}
    & .729 & .774 & .872 & .910 & .881 & .917
    & .703 & .821 & .733 & .838 & .785 & .870
    \\
    21 & VST$_{\textit{2021}}$ \cite{liu2021visual}
    & .778 & .813 & .897 & .928 & .910 & .932
    & .755 & .850 & .734 & .838 & .827 & .896
    \\ \rowcolor{mycolor}
    22 & DCENet$_{\textit{2021}}$ \cite{mei2021exploring}
    & .772 & .797 & .898 & .915 & .913 & .921
    & .754 & .839 & .731 & .835 & .832 & .882
    \\
    23 & PoolNet+$_{\textit{2022}}$ \cite{liu2022poolnet+}
    & .731 & .781 & .865 & .908 & .880 & .915
    & .710 & .829 & .724 & .839 & .783 & .875
    \\  \rowcolor{mycolor}
    24 & DNA$_{\textit{2022}}$ \cite{liu2021dna}
    & .743 & .786 & .864 & .905 & .883 & .915
    & .696 & .818 & .729 & .833 & .762 & 858
    \\
    25 & ICON$_{\textit{2022}}$ \cite{zhuge2022salient}
    & \textbf{.794} & \textbf{.817} & .902 & .920 & .918 & 929
    &  .761 & .844 & .752 & .844 & .848 & .901
    \\ \rowcolor{mycolor}
    26 & RCSBNet$_{\textit{2022}}$ \cite{ke2022recursive}
    & .789 & .813 & .909 & .919 & .916 & .922
    & .752 & .835 & .749 & .841 & .839 & .881
    \\
    27 & \textbf{ABiU-Net (ours)} & .785 & .797
    & \textbf{.928} & \textbf{.932} & \textbf{.935} & \textbf{.936}
    & \textbf{.800} & \textbf{.860} & \textbf{.778} & \textbf{.854}
    & \textbf{.873} & \textbf{.904}
    \\ \hline
\end{tabular}}
\end{table*}

\section{Experiments} \label{sec:experiments}
\subsection{Experimental Setup} \label{ssec:setup}
\subsubsection{Implementation Details}
We adopt the PyTorch framework \cite{paszke2019pytorch}
to implement the proposed method.
The backbone network, \ie, PVT \cite{wang2021pyramid},
is pre-trained on the ImageNet dataset \cite{russakovsky2015imagenet}.
The AdamW \cite{loshchilov2019decoupled} optimizer with 
the weight decay of 1e-4 is used to optimize the network.
The learning rate policy is \textit{poly} so that the current
learning rate equals the base one multiplying
$(1-curr\_iter/max\_iter)^{power}$,
where $curr\_iter$ and $max\_iter$ mean the numbers of 
the current and maximum iterations, respectively.
We set the initial learning rate to 5e-5 and $power$ to 0.9.
The proposed ABiU-Net is trained for 50 epochs 
with a batch size of 16.
All experiments are conducted on a TITAN Xp GPU.

\subsubsection{Datasets}
We follow recent studies
\cite{wang2018detect,liu2018picanet,wang2017stagewise,liu2021dna,liu2022poolnet+,qiu2019revisiting,zhao2019pyramid,wu2019cascaded,feng2019attentive,zeng2019towards,sun2021ampnet}
to train the proposed ABiU-Net on the DUTS training set 
\cite{wang2017learning}.
The DUTS training set is comprised of 10553 images and 
corresponding high-quality saliency map annotations. 
To evaluate the performance of various SOD methods,
we utilize the DUTS test set \cite{wang2017learning} and
five other widely used datasets, including
SOD \cite{movahedi2010design},
HKU-IS \cite{li2015visual},
ECSSD \cite{yan2013hierarchical},
DUT-OMRON \cite{yang2013saliency}, 
and THUR15K \cite{cheng2014salientshape}.
There are 5019, 300, 4447, 1000, 5168, and 6232 natural complex 
images in the above six test datasets, respectively.

\subsubsection{Evaluation Criteria}
This paper evaluates the accuracy of various SOD models using
four popular evaluation metrics, including 
the max $F$-measure score $F_\beta$, 
mean absolute error (MAE),
weighted $F$-measure score $F_\beta^\omega$ \cite{margolin2014evaluate},
and structure-measure $S_m$ \cite{fan2017structure}.
Here, the performance on a dataset is the average 
of all images in this dataset.
For the metrics of $F_\beta$, MAE, and $F_\beta^\omega$,
we use the same evaluation code as \cite{liu2020lightweight,liu2021dna,liu2021samnet,wu2022edn,qiu2022a2sppnet}.
For the metric of $S_m$, we adopt the official evaluation code \cite{fan2017structure}.
We introduce these metrics as follows.

Suppose $F_\beta$ denotes the $F$-measure score.
It is a weighted harmonic mean of precision and recall.
Specifically, given a threshold in the range of $[0, 1]$, 
the predicted saliency map can be converted into a binary map
that is compared to the ground-truth saliency map for computing
the precision and recall values.
Varying the threshold values, we can derive a series of 
precision-recall value pairs.
With precision and recall, $F_\beta$ can be formulated as
\begin{equation} \label{equ:Fb}
F_\beta = \frac{(1+\beta^2) \times {\rm Precision} \times {\rm Recall}}{
\beta^2 \times {\rm Precision} + {\rm Recall}},
\end{equation}
in which $\beta^2$ is set to a typical value of 0.3 to 
emphasize more on precision, following previous works
\cite{wang2015deep,lee2016deep,wang2016saliency,li2016deep,zhang2017amulet,zhang2017learning,luo2017non,wang2017stagewise,liu2018picanet,li2018contour,chen2018reverse,liu2020lightweight,liu2021samnet,qiu2019revisiting,qiu2020simple,wu2019cascaded,liu2022poolnet+,zhao2019egnet,liu2021dna,hou2019deeply,wang2018detect,zhang2018progressive,zhao2019pyramid,feng2019attentive,zeng2019towards,zeng2018learning}.
We compute $F_\beta$ scores under various thresholds and report 
the best one, \ie, the maximum $F_\beta$.

The MAE metric measures the absolute error between the predicted 
saliency map and the ground-truth map.
MAE is calculated as 
\begin{equation}
{\rm MAE} = \frac{1}{H \times W} \sum_{i=1}^{H} \sum_{j=1}^{W} |\mathbf{P}(i,j)-\mathbf{G}(i,j)|,
\end{equation}
in which $\mathbf{P}$ and $\mathbf{G}$ denote the predicted and 
ground-truth saliency maps, respectively.
$H$ and $W$ are image height and width, respectively.
$\mathbf{P}(i,j)$ and $\mathbf{G}(i,j)$ are the saliency scores
of the predicted and ground-truth saliency maps at the
location $(i,j)$, respectively.

We continue by introducing the weighted $F$-measure score
\cite{margolin2014evaluate}, denoted as $F_\beta^\omega$.
It is computed as 
\begin{equation}
F_\beta^\omega = \frac{(1+\beta^2) \times {\rm Precision}^\omega \times {\rm Recall}^\omega}{
\beta^2 \times {\rm Precision}^\omega + {\rm Recall}^\omega},
\end{equation}
where ${\rm Precision}^\omega$ and ${\rm Recall}^\omega$
are the weighted precision and weighted recall to amend
the flaws in other metrics.
The term $\beta^2$ has the same meaning as that in \equref{equ:Fb}.
Please refer to \cite{margolin2014evaluate} for more details.

Considering that the above measures are based on pixel-wise errors and often ignore the structural similarities, structure-measure ($S_m$) \cite{fan2017structure} is proposed to simultaneously evaluate region-aware and object-aware structural similarities.
$S_m$ is calculated as 
\begin{equation}
S_m = (1-\gamma) S_r + \gamma S_o,
\end{equation}
where $S_r$ and $S_o$ are region-aware and object-aware structural similarities, respectively. 
The balance parameter $\gamma$ is set to 0.5 by default.
We refer readers to the original paper \cite{fan2017structure} 
for the calculation of $S_r$ and $S_o$.

\begin{figure}[!tb]
    \centering
    \includegraphics[width=.95\linewidth]{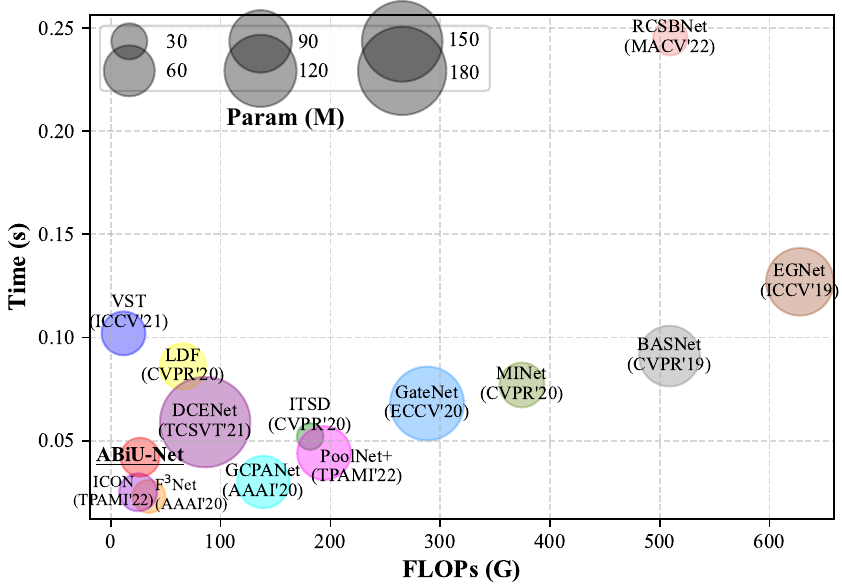}
    \\ \vspace{-0.05in}
    \caption{Visual comparison of ABiU-Net with some recent competitive methods in terms of parameters, FLOPs, and runtime.}
    \label{fig:complexity}
\end{figure}

\begin{figure}[!t]
\centering
\includegraphics[width=\linewidth]{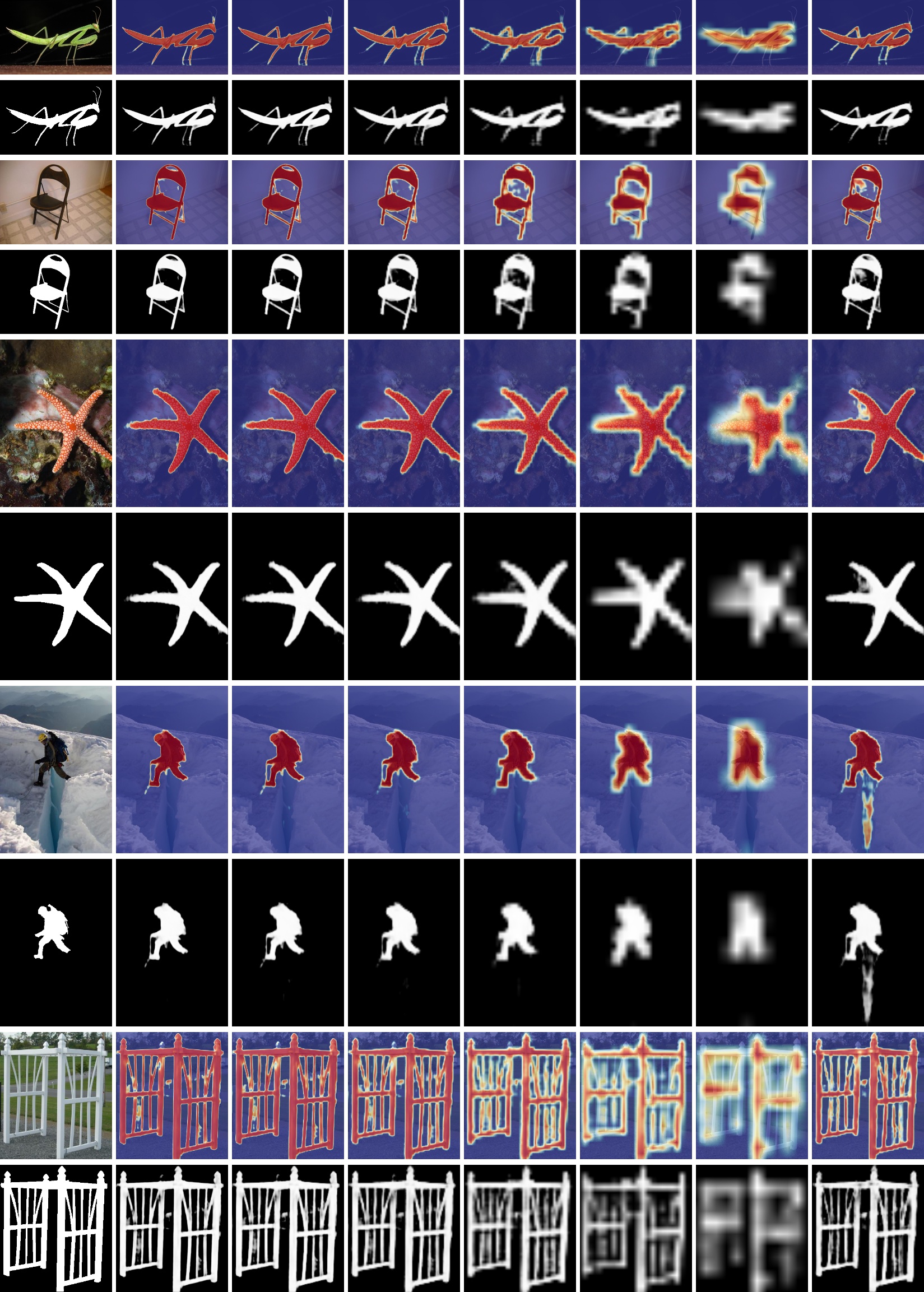}
    \\ \vspace{-0.05in}
    \leftline{\scriptsize
    \hspace{-0.04in} Img \& GT
    \hspace{0.09in} $\mathbf{P}^H_1$
    \hspace{0.2in} $\mathbf{P}^H_2$
    \hspace{0.2in} $\mathbf{P}^H_3$
    \hspace{0.2in} $\mathbf{P}^H_4$
    \hspace{0.2in} $\mathbf{P}^H_5$
    \hspace{0.2in} $\mathbf{P}^H_6$
    \hspace{0.2in} $\mathbf{P}^T$}
\caption{Feature visualization maps and saliency maps of
various intermediate side-outputs of ABiU-Net. 
Note that $\mathbf{P}^H_1$ is the final output saliency map.}
\label{fig:intermediate}
\end{figure}

\begin{figure*}[!tb]
    \centering
    \newcommand{\vsp}{\vspace{0.03in}}
    \leftline{\scriptsize Complex Scenes} \vsp
    \includegraphics[width=\linewidth]{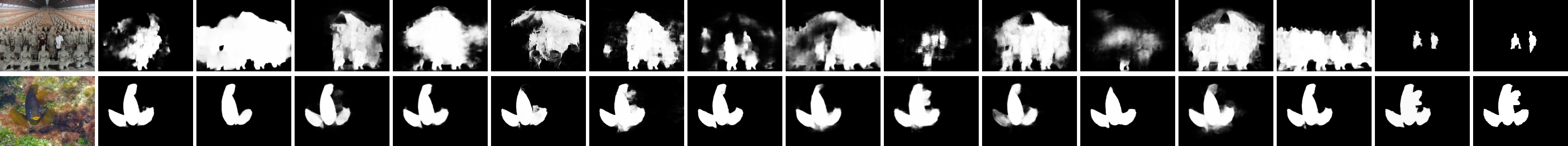}
    \\ \vspace{-0.16in} \rule[-1.5mm]{\linewidth}{0.2mm} 
    \leftline{\scriptsize Large Objects} \\ \vsp
    \includegraphics[width=\linewidth]{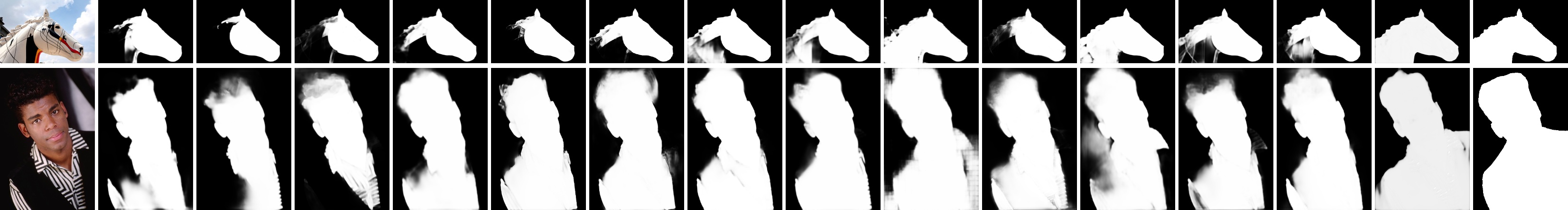}
    \\ \vspace{-0.16in} \rule[-1.5mm]{\linewidth}{0.2mm} 
    \leftline{\scriptsize Small Objects} \\ \vsp
    \includegraphics[width=\linewidth]{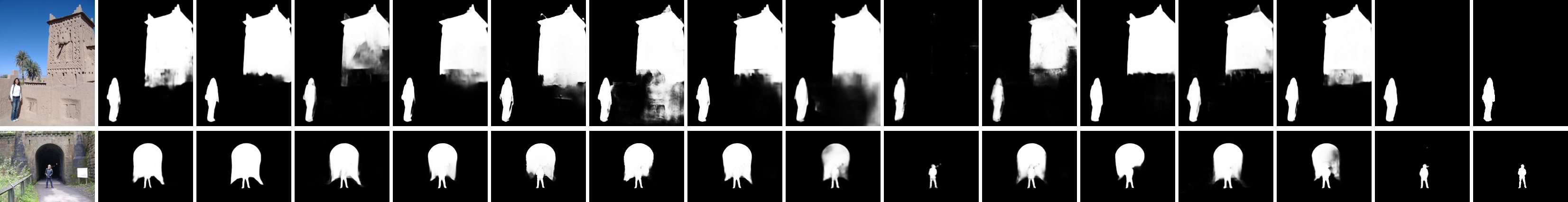}
    \\ \vspace{-0.16in} \rule[-1.5mm]{\linewidth}{0.2mm} 
    \leftline{\scriptsize Thin Objects} \\ \vsp
    \includegraphics[width=\linewidth]{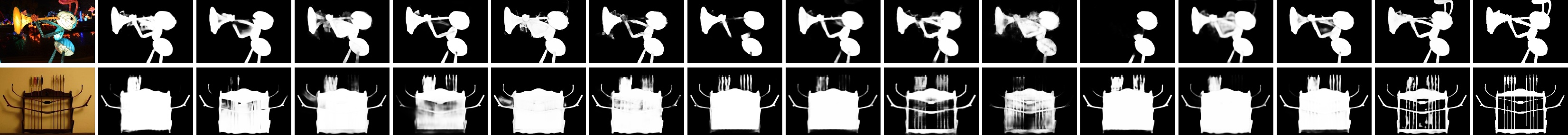} 
    \\ \vspace{-0.16in} \rule[-1.5mm]{\linewidth}{0.2mm} 
    \leftline{\scriptsize Multiple Objects} \\ \vsp
    \includegraphics[width=\linewidth]{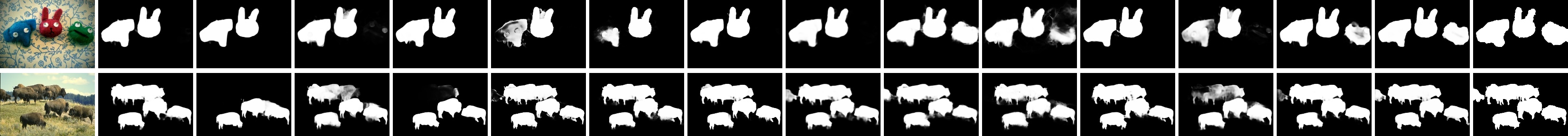}
    \\ \vspace{-0.16in} \rule[-1.5mm]{\linewidth}{0.2mm} 
    \leftline{\scriptsize Low Contrast} \\ \vsp
    \includegraphics[width=\linewidth]{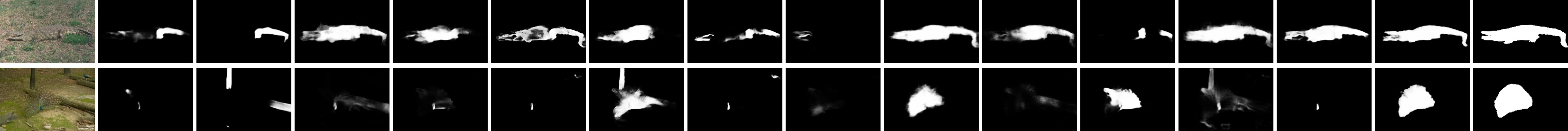}
    \\ \vspace{-0.16in} \rule[-1.5mm]{\linewidth}{0.2mm} 
    \leftline{\scriptsize Confusing Backgrounds} \\ \vsp
    \includegraphics[width=\linewidth]{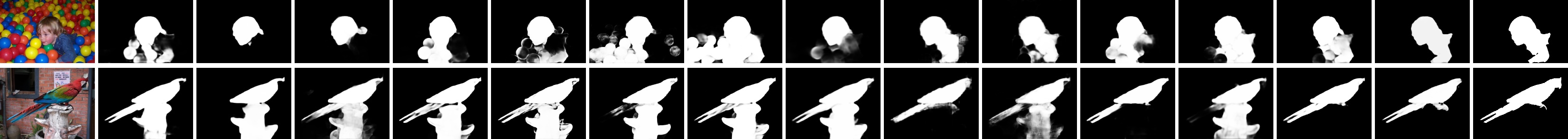}
    \\ \vspace{-0.16in} \rule[-1.5mm]{\linewidth}{0.2mm} 
    \leftline{\scriptsize Natural Phenomena} \\ \vsp
    \includegraphics[width=\linewidth]{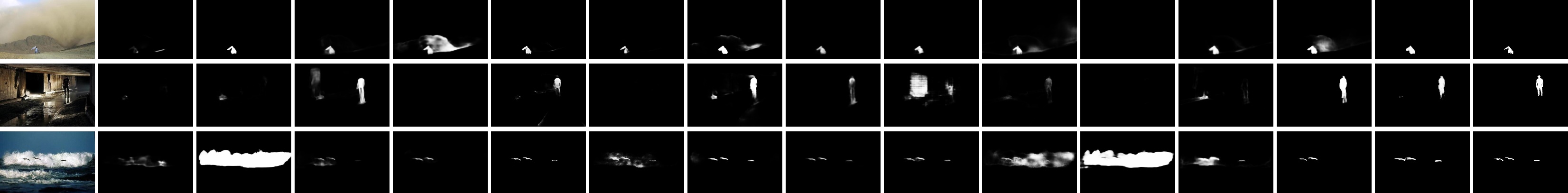}
    \\ \vspace{-0.05in}
    \leftline{\scriptsize\hspace{0.05in} Image 
    \hspace{0.18in} CPD \hspace{0.12in} BASNet
    \hspace{0.07in} EGNet \hspace{0.11in} F3Net
    \hspace{0.16in} ITSD \hspace{0.15in} MINet
    \hspace{0.16in} LDF \hspace{0.1in} GCPANet
    \hspace{0.08in} VST \hspace{0.12in} PoolNet+
    \hspace{0.03in} DCENet \hspace{0.10in} DNA
    \hspace{0.15in} ICON \hspace{0.16in} Ours
    \hspace{0.20in} GT}
    \vspace{-1.mm}
    \caption{Qualitative comparison between ABiU-Net and 13 \sArt SOD methods.}
    \vspace{-1.2mm}
    \label{fig:samples}
\end{figure*}

\subsection{Performance Comparison}
We compare the proposed ABiU-Net to previous
state-of-the-art SOD methods, including 
UCF \cite{zhang2017learning}, SRM \cite{wang2017stagewise},
PiCA \cite{liu2018picanet}, BRN \cite{wang2018detect},
C2S \cite{li2018contour}, RAS \cite{chen2018reverse}, 
DSS \cite{hou2019deeply}, PAGE-Net \cite{wang2019salient}, 
AFNet \cite{feng2019attentive}, DUCRF \cite{xu2019structured}, 
HRSOD \cite{zeng2019towards}, CPD \cite{wu2019cascaded},
BASNet \cite{qin2019basnet}, EGNet \cite{zhao2019egnet}, 
F$^3$Net \cite{wei2020f3net}, ITSD \cite{zhou2020interactive}, 
MINet \cite{pang2020multi}, LDF \cite{wei2020label}, 
GCPANet \cite{chen2020global}, GateNet \cite{zhao2020suppress}, 
VST \cite{liu2021visual}, DCENet \cite{mei2021exploring}, 
PoolNet+ \cite{liu2022poolnet+}, DNA \cite{liu2021dna}, 
ICON \cite{zhuge2022salient}, and RCSBNet \cite{ke2022recursive}.
For fair comparisons, the predicted saliency maps are downloaded
from the official websites or produced by the released code
with default settings.
Note that we do not provide the results of MDF \cite{li2015visual} 
on the HKU-IS \cite{li2015visual} dataset because MDF adopts 
HKU-IS for training.
For the same reason, we do not report the results of DHS \cite{liu2016dhsnet} on the DUT-OMRON \cite{yang2013saliency} dataset.

\subsubsection{Quantitative Evaluation}
The quantitative results of various methods in terms of $F_\beta$, MAE, $F_\beta^\omega$, and $S_m$ on six datasets are summarized in \tabref{tab:eval_f_mae}.
We can observe that the proposed ABiU-Net outperforms other methods by a large margin.
Specifically, ABiU-Net attains $F_\beta$ values of 87.9\%, 95.1\%, 95.9\%, 84.3\%, 82.0\%, and 90.6\%, which are
0\%, 0.9\%, 0.8\%, 2.1\%, 0.3\%, and 1.6\% higher than the second-best results on SOD, HKU-IS, ECSSD, DUT-OMRON, THUR15K, and DUTS-test datasets, respectively.
ABiU-Net also achieves the best MAE, \ie, 0.6\%, 0.4\%, 0.6\%, 0.5\%, and 0.6\% better than the second-best results on HKU-IS, ECSSD, DUT-OMRON, THUR15K, and DUTS-test datasets, respectively.
On the SOD dataset, the MAE value of ABiU-Net is a little worse than that of
VST \cite{liu2021visual}, ICON \cite{zhuge2022salient},
and RCSBNet \cite{ke2022recursive}.
Moreover, the $F_\beta^\omega$ values of ABiU-Net are 1.9\%, 1.7\%, 3.9\%, 2.6\%, 
and 2.5\% higher than the second-best results on HKU-IS, ECSSD, DUT-OMRON, THUR15K,
and DUTS-test datasets, respectively.
In terms of the metric $S_m$, ABiU-Net achieves the best performance in all datasets except for SOD.
The $S_m$ values of ABiU-Net are 0.4\%, 0.4\%, 1.0\%, 1.0\%, and 0.3\% higher than the second-best results
on HKU-IS, ECSSD, DUT-OMRON, THUR15K, and DUTS-test datasets, respectively.
Therefore, we can conclude that ABiU-Net has pushed forward the \sArt for SOD significantly.

\begin{figure*}[!t]
    \centering
    \includegraphics[width=.4\linewidth]{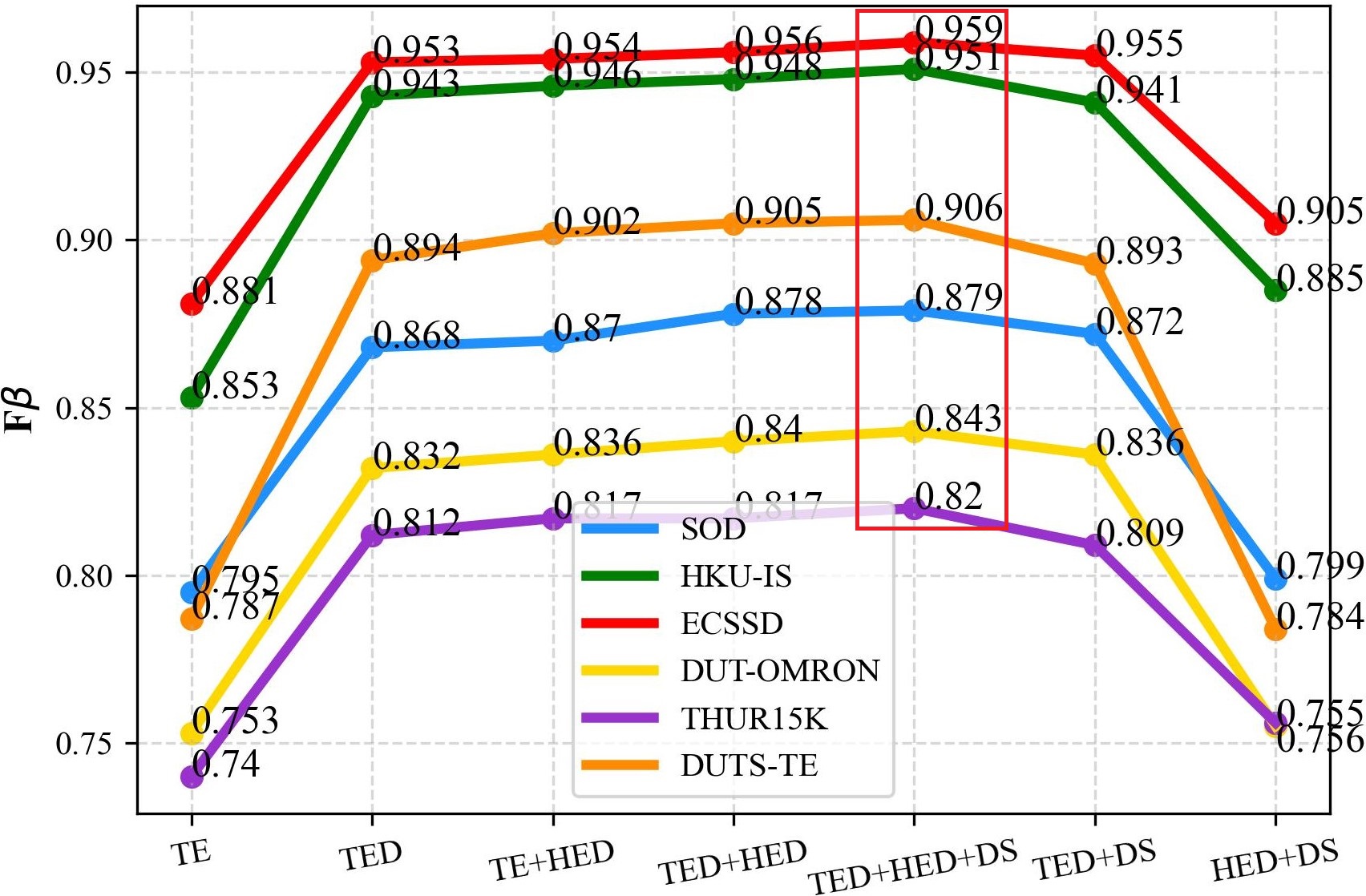}
    \hspace{0.5in}
    \includegraphics[width=.4\linewidth]{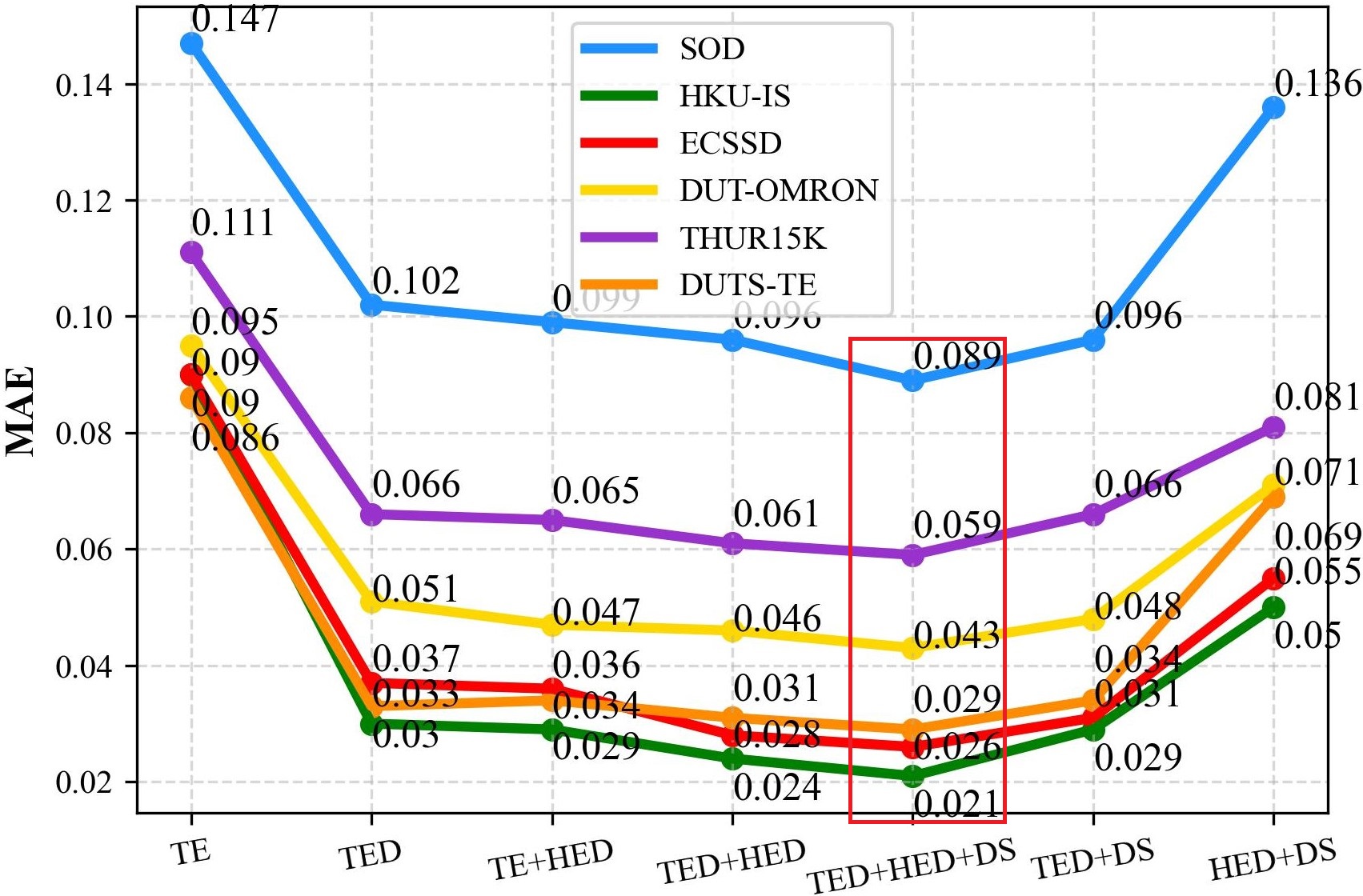}
    \\ \vspace{-0.05in}
    \caption{Effect of the main components of ABiU-Net.
    TE: TEncPath; TED: TEncPath + TDecPath; 
    HED: HEncPath + HDecPath; DS: Deep Supervision.}
    \label{fig:ablation_modules}
\end{figure*}

\begin{table*}[!t]
\centering
\setlength{\tabcolsep}{2.2mm}
\caption{Ablation studies for the deep supervision strategies of ABiU-Net.
} \label{tab:ablation_ds}
\resizebox{.95\linewidth}{!}{
\begin{threeparttable}%
\begin{tabular}{c||c|c|c|c|c|c|c|c|c|c|c|c} \hline
    \rowcolor{mycolor}
    & \multicolumn{2}{c|}{SOD} & \multicolumn{2}{c|}{HKU-IS}
    & \multicolumn{2}{c|}{ECSSD} & \multicolumn{2}{c|}{DUT-OMRON}
    & \multicolumn{2}{c|}{THUR15K} & \multicolumn{2}{c}{DUTS-test}
    \\ \cline{2-13} \rowcolor{mycolor}
    \multirow{-2}{*}{Deep Supervision}
    & $F_\beta$ & MAE & $F_\beta$ & MAE & $F_\beta$ & MAE 
    & $F_\beta$ & MAE & $F_\beta$ & MAE & $F_\beta$ & MAE 
    \\ \hline
    \{$\mathbf{P}^H_1$, $\mathbf{P}^T_1$\}
    & .878 & .096 & .948 & .024 & .956 & .028
    & .840 & .046 & .817 & .061 & .905 & .031
    \\ \rowcolor{mycolor}
    \{$\mathbf{P}^H_i$($i\in \{1,2,3,4,5,6\}$), $\mathbf{P}^T_1$\}
    & \textbf{.879} & \textbf{.089} & \textbf{.951} & \textbf{.021}
    & .959 & \textbf{.026} & \textbf{.843} & \textbf{.043}
    & \textbf{.820} & \textbf{.059} & \textbf{.906} & \textbf{.029}
    \\
    \{$\mathbf{P}^H_i$, $\mathbf{P}^T_i$($i\in \{1,2,3,4\}$)\}
    & .875 & .091 & .948 & .023 & .959 & .027
    & .841 & \textbf{.043} & .817 & .062 & \textbf{.906} & \textbf{.029}
    \\ 
    \{$\mathbf{P}^H_i$($i\in \{1,2,3,4,5,6\}$), $\mathbf{P}^T_i$($i\in \{1,2,3,4\}$)\}
    & .876 & .092 & .948 & .024 & \textbf{.960} & .027
    & .842 & \textbf{.043} & \textbf{.820} & .062 & .905 & .030
    \\ \hline
\end{tabular}
\begin{tablenotes}
\item[*] ``\{$\mathbf{P}^H_i$($i\in \{1,2,3,4,5,6\}$), $\mathbf{P}^T_1$\}''
means that the output saliency maps of all stages of HDecPath and the output saliency map of the last stage of TDecPath (\ie, $\mathbf{F}^{\rm TD}_1$) are supervised by ground-truth, which is the default deep supervision strategy of our ABiU-Net.
\end{tablenotes}
\end{threeparttable}}
\end{table*}

\subsubsection{Complexity Analysis}
\figref{fig:complexity} displays the complexity comparison between ABiU-Net and some recent competitive methods in terms of the number of parameters, the number of FLOPs, and runtime.
Note that we use the same hardware setup (\ie, TITAN Xp GPU) and framework (\ie, PyTorch \cite{paszke2019pytorch}) to test all methods for complexity analysis.
It can be observed that the complexity of ABiU-Net is comparable to recent counterparts, exhibiting a relatively small number of parameters and FLOPs, as well as a relatively short runtime.

\subsubsection{Qualitative Evaluation}
In \figref{fig:intermediate}, we display some feature
visualization figures and the corresponding saliency maps of various intermediate side-outputs to show how features evolve in ABiU-Net.
From \figref{fig:intermediate}, it can be seen that the shapes of
salient objects can be markedly refined from $\mathbf{P}^H_6$ to
$\mathbf{P}^H_1$, which suggests the importance of the multi-level 
decoder in SOD. Moreover, the quality of saliency maps of 
$\mathbf{P}^H_1$ is far better than that of $\mathbf{P}^T$, 
which demonstrates the effectiveness of ABiU-Net in segmenting 
more accurate objects by learning complementary global and local
information.

We proceed by displaying the qualitative comparisons in \figref{fig:samples} 
to explicitly show the superiority of ABiU-Net over previous
\sArt SOD methods.
\figref{fig:samples} includes some representative images
to incorporate various difficult circumstances, including 
complex scenes, large/small objects, thin objects,
multiple objects, low-contrast scenes, and confusing backgrounds.
Overall, ABiU-Net can generate better saliency maps
in various scenarios.
Surprisingly, ABiU-Net can even accurately segment
salient objects with very complicated thin structures
(the second thin sample), which is very challenging
for all other methods.
In addition, we also provide some representative images in the last group of \figref{fig:samples} to show the results of ABiU-Net when being applied to natural
phenomena like clouds, smoke, wave, and strong illumination at night. 
Overall, ABiU-Net can accurately segment salient objects under various natural phenomena. 
Overall, such a good visual performance of ABiU-Net benefits from its bilateral
encoder/decoder structure. The communication between the two encoder/decoder paths
facilitates ABiU-Net to accurately segment salient objects with fine details in
various scenes, which is the core reason why our ABiU-Net can achieve such 
a good visual performance. In detail, the TEncPath/TDecPath of ABiU-Net can determine the coarse
locations and shapes of salient objects, which is essential to find the large objects,
multiple objects, and objects in complex, low-contrast or confusing background scenes.
The HEncPath/HDecPath of ABiU-Net can model the local fine-grained representations to further
refine salient object details guided by the coarse locations and shapes modeled
by TEncPath/TDecPath, which is helpful to segment small/thin objects and
distinguish the boundaries of salient objects.

\begin{table*}[!t]
\centering
\setlength{\tabcolsep}{1.8mm}
\caption{Ablation studies for the hyper-parameters of ABiU-Net.
} \label{tab:ablation_paras}
\resizebox{.95\linewidth}{!}{
\begin{threeparttable}%
\begin{tabular}{l|c||c|c|c|c|c|c|c|c|c|c|c|c} \hline
    \rowcolor{mycolor} \multicolumn{2}{c||}{} 
    & \multicolumn{2}{c|}{SOD} & \multicolumn{2}{c|}{HKU-IS}
    & \multicolumn{2}{c|}{ECSSD} & \multicolumn{2}{c|}{DUT-OMRON}
    & \multicolumn{2}{c|}{THUR15K} & \multicolumn{2}{c}{DUTS-test}
    \\ \cline{3-14} \rowcolor{mycolor}
    \multicolumn{2}{c||}{\multirow{-2}{*}{Configurations}}
    & $F_\beta$ & MAE & $F_\beta$ & MAE & $F_\beta$ & MAE 
    & $F_\beta$ & MAE & $F_\beta$ & MAE & $F_\beta$ & MAE 
    \\ \hline
    \multicolumn{2}{c||}{Default Configurations}
    & .879 & \textbf{.089} & \textbf{.951} & \textbf{.021}
    & \textbf{.959} & .026 & .843 & \textbf{.043}
    & \textbf{.820} & \textbf{.059} & .906 & .029
    \\ \hline
    \multirow{4}*{{\#Channels of HEncPath}}
    & (128, 128, 64, 32, 32, 8)
    & .879 & .092 & \textbf{.951} & .022 & .958 & .027
    & .839 & .047 & .814 & .062 & .901 & .030
    \\
    & (256, 128, 64, 32, 16, 8)
    & .879 & .093 & \textbf{.951} & .022 & \textbf{.959} & .026
    & .843 & .045 & .815 & .061 & .904 & .029
    \\
    & (512, 256, 128, 64, 32, 16)
    & .877 & .093 & \textbf{.951} & .022 & .958 & .027
    & .839 & .047 & .815 & .062 & .904 & .030
    \\
    & (512, 256, 128, 128, 64, 16)
    & \textbf{.881} & .090 & .950 & .022 & \textbf{.959} 
    & \textbf{.025} & .836 & .048 & .814 & .062 & .901 & .030
    \\ \hline
    \multirow{4}*{{\#Channels of TDecPath}}
    & (64, 32, 16, 8)
    & .878 & .093 & \textbf{.951} & .022 & .957 & .027
    & .844 & .045 & .817 & .060 & \textbf{.907} & \textbf{.028}
    \\
    & (128, 32, 32, 8)
    & .868 & .096 & \textbf{.951} & .022 & .957 & .028
    & .841 & .045 & .818 & .060 & \textbf{.907} & \textbf{.028}
    \\
    & (128, 128, 64, 16)
    & .876 & .092 & \textbf{.951} & .022 & .958 & .027
    & .842 & .045 & .818 & .060 & \textbf{.907} & \textbf{.028}
    \\
    & (256, 128, 64, 32)
    & .871 & .097 & \textbf{.951} & .022 & \textbf{.959} & .027
    & .844 & .045 & .817 & .060 & \textbf{.907} & .029
    \\ \hline
    \multirow{4}*{{\#Channels of HDecPath}}
    & (128, 64, 32, 16, 16, 8)
    & .875 & .094 & .950 & .022 & \textbf{.959} & .026
    & .843 & .046 & .816 & .061 & .905 & .029
    \\
    & (128, 128, 64, 32, 16, 8)
    & .880 & .090 & \textbf{.951} & .022 & .958 & \textbf{.025}
    & .840 & .045 & .815 & .062 & .906 & .029
    \\
    & (256, 256, 64, 32, 16, 8)
    & .874 & .095 & \textbf{.951} & .022 & .958 & \textbf{.025}
    & \textbf{.846} & .045 & .813 & .062 & .904 & .029
    \\ 
    & (512, 256, 128, 64, 32, 16)
    & .872 & .096 & .950 & .022 & \textbf{.959} & .026
    & .840 & .045 & .815 & .061 & .906 & .029
    \\ \hline
\end{tabular}
\begin{tablenotes}
\item[*] ``\#Channels'' means the number of channels.
The default numbers of channels for HEncPath, TDecPath, and 
HDecPath are (256, 256, 128, 64, 64, 16), (128, 64, 32, 16),
and (256, 128, 64, 32, 32, 8) from top to bottom, respectively.
\end{tablenotes}
\end{threeparttable}}
\end{table*}

\subsection{Ablation Studies}
In this section, we conduct extensive ablation studies
for a better understanding of the proposed method.

\subsubsection{Effect of Component Designs}
We first evaluate the effect of the component designs of
the proposed ABiU-Net.
We start with the simple transformer-based encoder network
without the decoder, \ie, TEncPath.
We directly upsample the feature map from the last stage 
of the encoder for final prediction.
The results are shown in the first column of
\figref{fig:ablation_modules} (\ie, TE).
As can be seen, the performance is quite poor.

\paragraph{Effect of the Decoder}
Then, we add a simple decoder, \ie, TDecPath (with the attention mechanism),
to TEncPath, resulting in a transformer-based U-shaped encoder-decoder network.
The evaluation results are put in the second column of
\figref{fig:ablation_modules} (\ie, TED).
Significant performance boosting can be observed,
which demonstrates that the encoder-decoder structure is 
necessary to utilize the low-level features for accurate SOD.
The performance is even very competitive with recent \sArt
SOD methods, implying that the powerful global relationship
modeling of vision transformers 
\cite{vaswani2017attention,dosovitskiy2021image,li2021dense}
is essential for SOD in discovering salient objects.
The goal of this paper is to further boost the SOD performance
upon this high baseline by combing the merits of vision 
transformers and CNNs.

\paragraph{Effect of the Asymmetric Bilateral Encoder}
\label{sec:ablation_enc}
Next, we adopt TEncPath and HEncPath to build an asymmetric 
bilateral encoder.
We also adopt HDecPath as the decoder,
but the input of each stage in HDecPath is the outputs 
from HEncPath and the preceding decoder stage, excluding
the output from TDecPath in ABiU-Net.
The experimental results are displayed in the third
column of \figref{fig:ablation_modules} (\ie, TE+HED).
We can observe a significant performance improvement.
This experiment validates that our asymmetric bilateral 
encoder can supplement the transformer by complementary 
local representations, leading to higher SOD accuracy.

\paragraph{Effect of the Asymmetric Bilateral Decoder}
We continue by adding TDecPath (with attention mechanism) to the model
in \secref{sec:ablation_enc} to form the asymmetric bilateral decoder.
The evaluation results are summarized in the fourth column
of \figref{fig:ablation_modules} (\ie, TED+HED).
The asymmetric bilateral decoder can consistently improve 
the SOD accuracy.
This experiment verifies that the asymmetric bilateral decoder 
is helpful in learning complementary information for both 
decoding the locations and fine details of salient objects.

\paragraph{Effect of Deep Supervision}
At last, we add deep supervision as in \figref{fig:frame}
to obtain the final ABiU-Net.
The results are shown in the fifth column
of \figref{fig:ablation_modules} (\ie, TED+HED+DS).
Such default ABiU-Net achieves the best accuracy.
Note that the fourth column is just ABiU-Net 
without deep supervision.
From the fourth column to the fifth column,
there is a significant performance improvement,
indicating the effectiveness of training with deep supervision.
Moreover, the default deep supervision strategy we used is to impose deep supervision on saliency maps of all stages of HDecPath (\ie, $\mathbf{P}^H_i$($i\in \{1,2,3,4,5,6\}$) while only on the saliency map of the last stage of TDecPath (\ie, $\mathbf{P}^T_1$). 
Intuitively, the reason for this design is that imposing deep supervision on all
saliency maps of HDecPath can provide a direct supervision for each hidden layer of HDecPath
(\ie, $\mathbf{F}^{\rm HD}_6$ $\sim$ $\mathbf{F}^{\rm HD}_1$). Moreover, this supervision can be
easily propagated back to hidden layers of TDecPath to achieve the implicit optimization of
all feature maps of TDecPath (\ie, $\mathbf{F}^{\rm TD}_4$ $\sim$ $\mathbf{F}^{\rm TD}_1$),
because $\mathbf{F}^{\rm TD}_4$ $\sim$ $\mathbf{F}^{\rm TD}_1$ are the direct inputs of HDecPath
to generate $\mathbf{F}^{\rm HD}_6$ $\sim$ $\mathbf{F}^{\rm HD}_3$. Hence, only supervising the
output saliency map of the last stage of TDecPath rather than the saliency maps of all stages is
enough to improve the discriminativeness of the hidden layer features of TDecPath.

In addition, we conduct the following ablation studies to further evaluate the effect of the different deep supervision strategies experimentally, whose results are shown in \tabref{tab:ablation_ds}. 
It is clear that all deep supervision strategies can boost the detection performance, and the model with the default deep supervision strategy can achieve the best overall performance.

\begin{figure}[!tb]
    \centering
    \includegraphics[width=.155\linewidth]{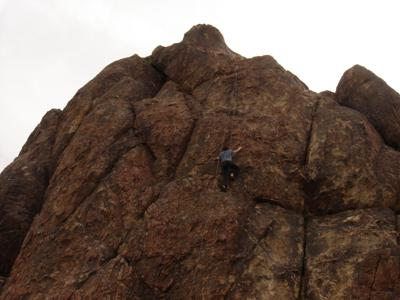}
    \includegraphics[width=.155\linewidth]{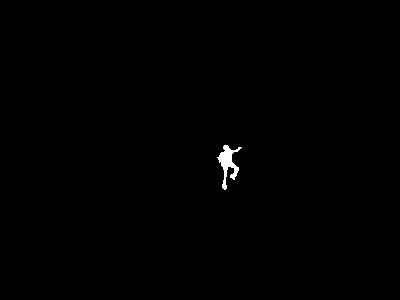}
    \includegraphics[width=.155\linewidth]{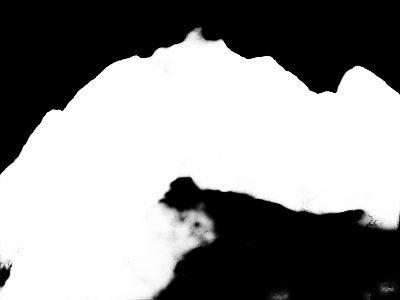}
    \includegraphics[width=.155\linewidth]{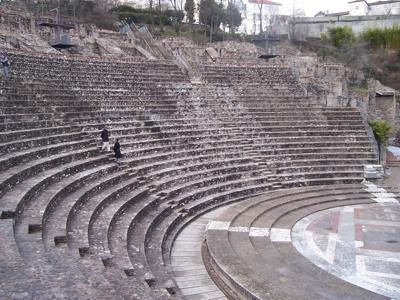}
    \includegraphics[width=.155\linewidth]{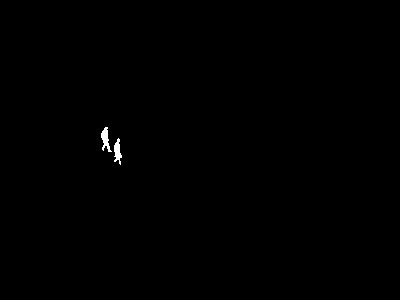}
    \includegraphics[width=.155\linewidth]{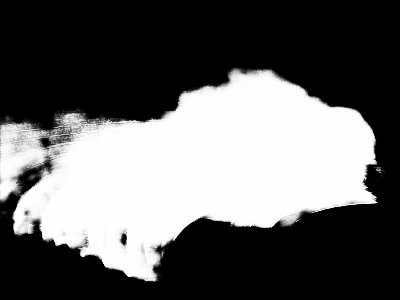}
    \\ \vspace{0.015in}
    \includegraphics[width=.155\linewidth,height=1cm]{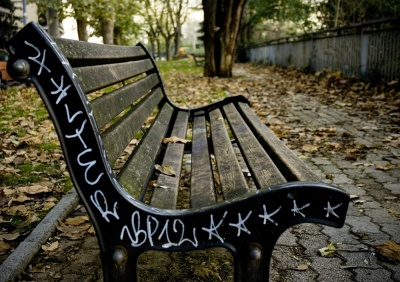}
    \includegraphics[width=.155\linewidth,height=1cm]{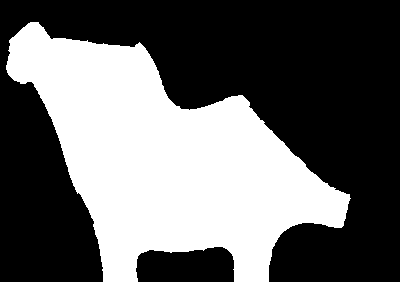}
    \includegraphics[width=.155\linewidth,height=1cm]{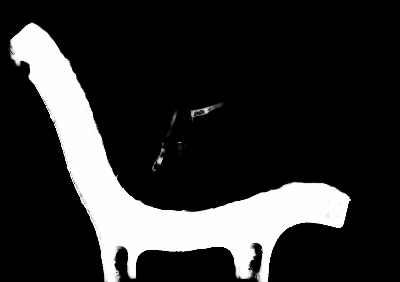}
    \includegraphics[width=.155\linewidth,height=1cm]{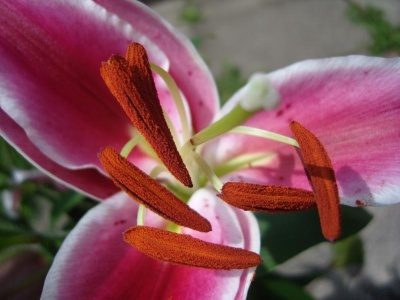}
    \includegraphics[width=.155\linewidth,height=1cm]{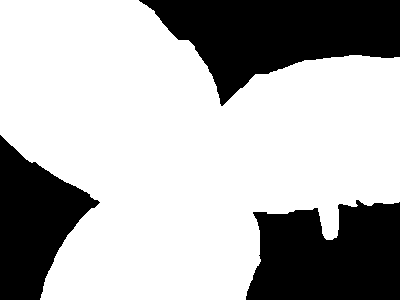}
    \includegraphics[width=.155\linewidth,height=1cm]{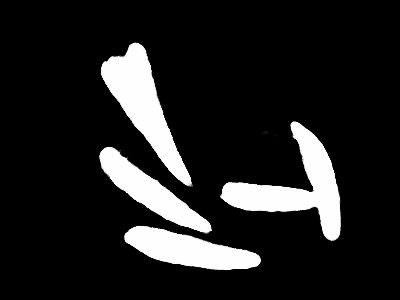}
    \\ \vspace{0.015in}
    \includegraphics[width=.155\linewidth,height=0.95cm]{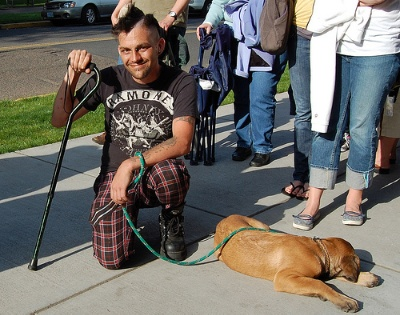}
    \includegraphics[width=.155\linewidth,height=0.95cm]{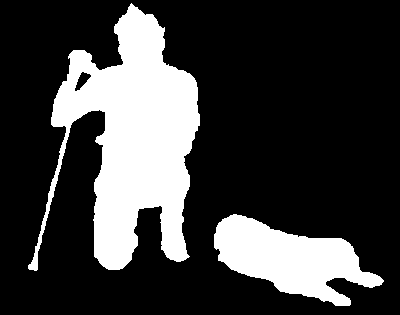}
    \includegraphics[width=.155\linewidth,height=0.95cm]{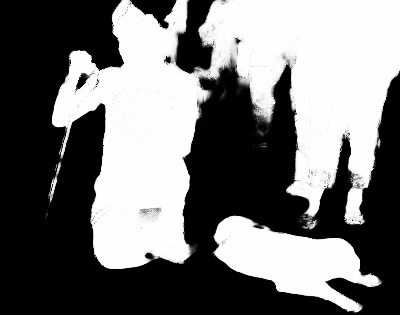}
    \includegraphics[width=.155\linewidth,height=0.95cm]{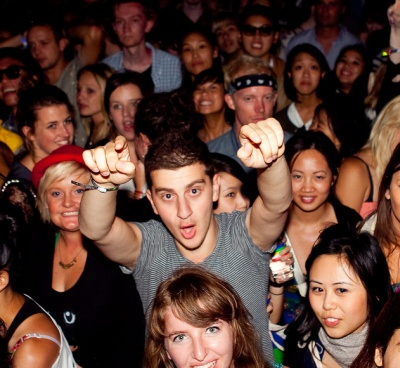}
    \includegraphics[width=.155\linewidth,height=0.95cm]{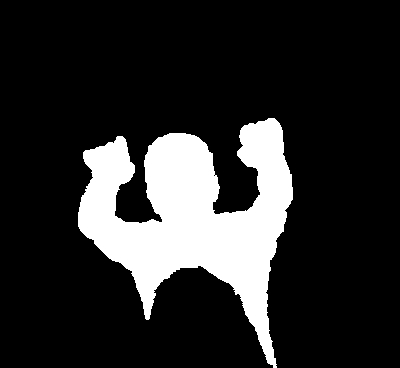}
    \includegraphics[width=.155\linewidth,height=0.95cm]{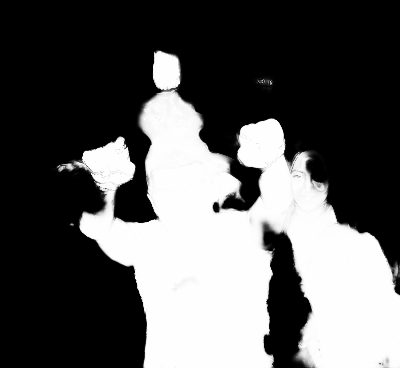}
    \\ \vspace{0.015in}
    \includegraphics[width=.155\linewidth]{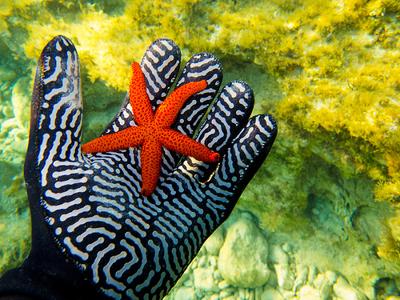}
    \includegraphics[width=.155\linewidth]{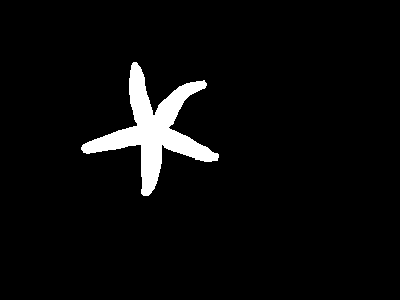}
    \includegraphics[width=.155\linewidth]{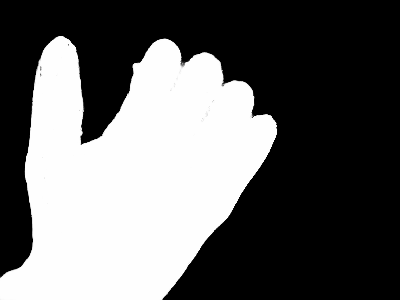}
    \includegraphics[width=.155\linewidth]{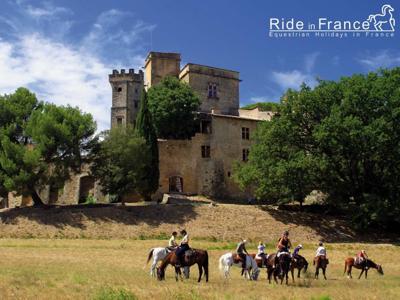}
    \includegraphics[width=.155\linewidth]{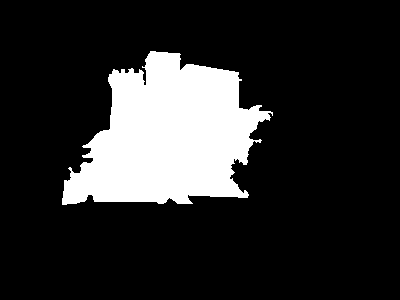}
    \includegraphics[width=.155\linewidth]{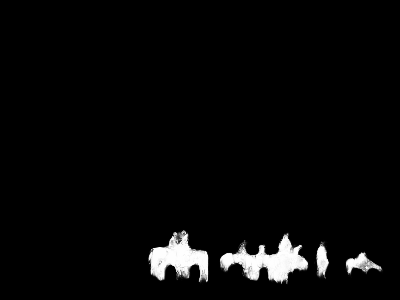}
    \\ \vspace{-0.05in}
    \leftline{\scriptsize
    \hspace{0.12in} Image \hspace{0.34in} GT \hspace{0.35in} Ours
    \hspace{0.32in} Image \hspace{0.31in} GT \hspace{0.38in} Ours}
    \caption{Some failure cases of the proposed ABiU-Net. GT: Ground Truth.}
    \label{fig:fail}
\end{figure}

\paragraph{Effect of Individual Transformer or CNN}
The superiority of ABiU-Net over previous \sArt methods is that ABiU-Net can make complementary use of the global contextual modeling ability of transformer and the local representation learning ability of CNNs.
Next, we conduct ablation studies to evaluate the effect when the individual transformer or CNN is considered separately, and the results are displayed in the sixth and seventh columns of \figref{fig:ablation_modules}. 
It is clear that the individual transformer or CNN with the CA and DS (\ie, TED+DS or HED+DS) would produce worse results than the default ABiU-Net (\ie, TED+HED+DS).
Hence, we can explicitly confirm the superiority of our asymmetric bilateral design. 
In addition, the CNN encoder path (\ie, HEncPath) is lightweight, so it can be optimized well with the large-scale SOD training data despite not being pre-trained on the ImageNet dataset \cite{he2019rethinking}.

\subsubsection{Impact of Hyper-parameters}
To explain how the default hyper-parameters of ABiU-Net are set,
we evaluate the performance when varying hyper-parameters.
Since PVT \cite{wang2021pyramid} is used as the TEncPath,
we study the numbers of channels of 
HEncPath ($\mathbf{F}^{\rm HE}_i$), 
TDecPath ($\hat{\mathbf{F}}^{\rm TE}_i$), 
and HDecPath ($\hat{\mathbf{F}}^{\rm HE}_i$).
We try some different settings, and the results are summarized 
in \tabref{tab:ablation_paras}.
One can observe that ABiU-Net seems quite robust to various parameter
settings as there is only a small performance fluctuation.
This property makes that ABiU-Net has the potential to 
serve as a base architecture for SOD in the transformer era.
Since the default parameter setting achieves slightly better
performance, we use this setting by default.

\subsection{Failure Case Analysis}
Although our ABiU-Net achieves a new state of the art, 
it may still fail in some cases which are reported 
in \figref{fig:fail}.
As can be seen from the two examples in the first row of
\figref{fig:fail}, our ABiU-Net may fail for tiny objects 
and less inconspicuous objects.
From the two examples in the second row, we can see that 
the partial regions within some saliency objects are particularly
prominent, resulting in the failure of our ABiU-Net which may only
segment the more prominent regions.
Hence, how to ensure the integrity of salient objects is 
the next problem to be solved. 
Moreover, as shown in the third row of \figref{fig:fail},
when the background regions are particularly complex, 
especially containing objects belonging to the same semantic categories as salient objects, our method may fail to 
distinguish the salient objects from background regions.
Furthermore, the salient objects in an image of the datasets are subjectively labeled and it may be ambiguous which objects are salient in a few images.
In this case, our ABiU-Net may detect other objects in an image to be salient, rather than the ground-truth salient objects, as shown in the fourth row of \figref{fig:fail}. 
Based on the above analysis, we can conclude that there is still a long way towards the ideal SOD solution.

\section{Conclusion}
This paper focuses on boosting SOD accuracy with the vision 
transformer \cite{vaswani2017attention,dosovitskiy2021image}.
It is widely accepted that vision transformers are adept at learning global contextual information that is essential for locating salient objects, while CNNs have a strong ability to learn local fine-grained information that is necessary for refining object details \cite{cheng2014global,luo2017non,liu2018picanet}.
Therefore, this paper explores the combination of the transformer and CNN to learn discriminative hybrid features for accurate SOD.
For this goal, we design the \textbf{A}symmetric \textbf{Bi}lateral \textbf{U-Net} (ABiU-Net), where both the encoder and decoder have two paths.
Extensive experiments demonstrate that ABiU-Net can significantly improve SOD performance when compared with \sArt methods.
Considering that ABiU-Net is an elegant architecture without carefully designed modules or engineering skills, ABiU-Net provides a new perspective for SOD in the transformer era.

{\small
\bibliographystyle{IEEEtran}
\bibliography{references}
}

\newcommand{\AddPhoto}[1]{\includegraphics[width=1in,clip,keepaspectratio]{#1}}

\begin{IEEEbiography}[\AddPhoto{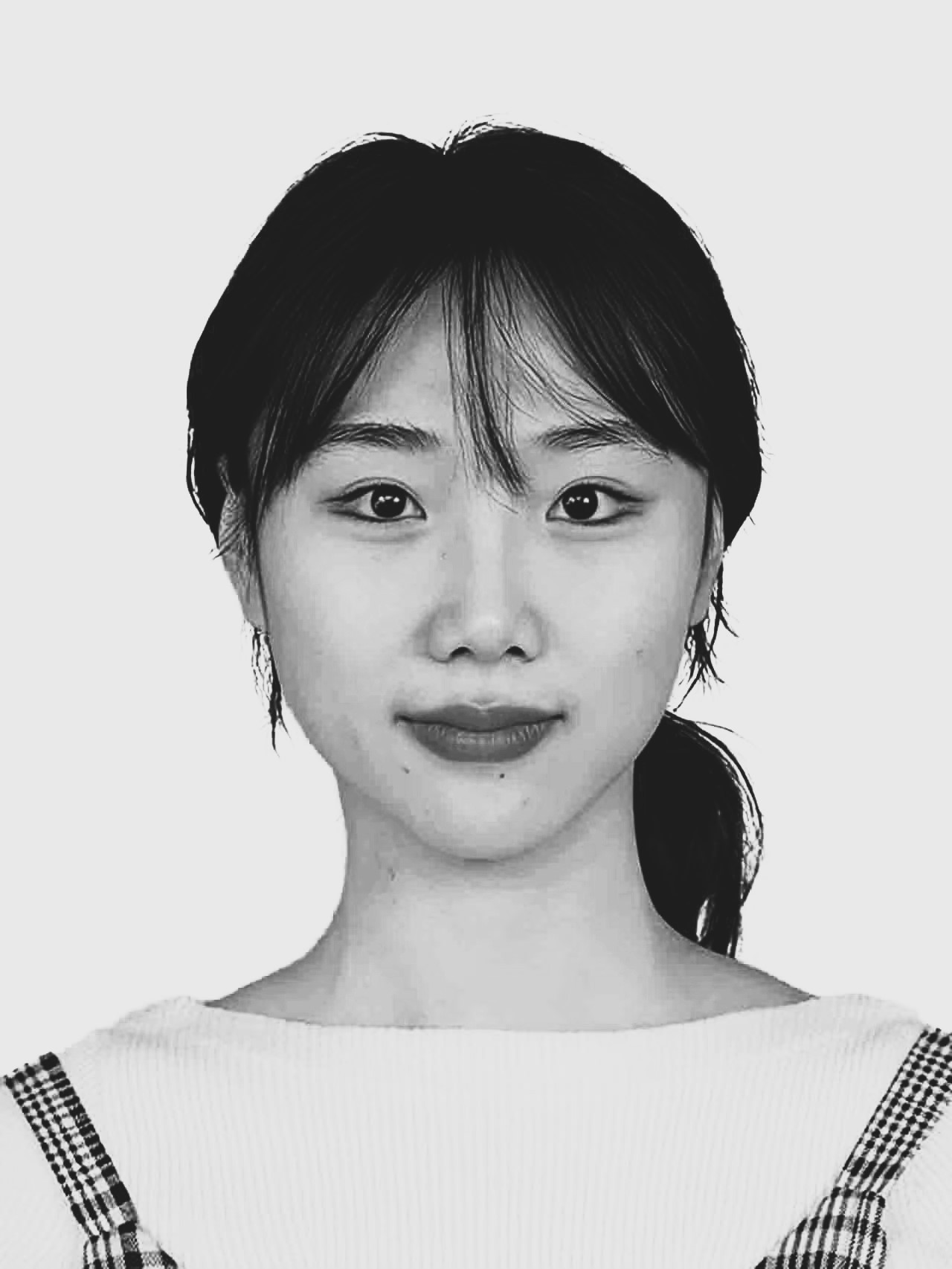}]{Yu Qiu}
received her BEng degree from Northwest A\&F University in 2017,
and her Ph.D. degree from Nankai University in 2022. Currently,
she is a postdoctoral scholar at the College of Artificial Intelligence, 
Nankai University. Her research interests include computer vision
and machine learning.
\end{IEEEbiography}

\begin{IEEEbiography}[\AddPhoto{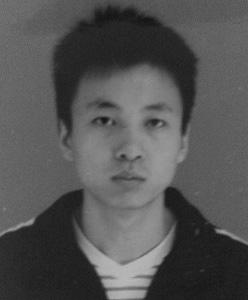}]{Yun Liu}
received his BEng and Ph.D. degrees from Nankai University in 2016 and 2020, respectively.
Then, he worked with Prof. Luc Van Gool for one and a half years as a postdoctoral scholar at Computer Vision Lab, ETH Zurich.
Currently, he is a scientist at Institute for Infocomm Research (I2R), A*STAR.
His research interests include computer vision and machine learning.
\end{IEEEbiography}

\begin{IEEEbiography}[\AddPhoto{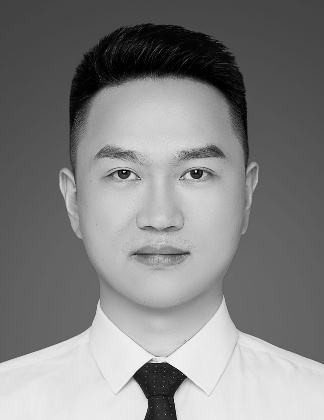}]{Le Zhang}
received the BEng degree from the University of Electronic Science and Technology of China, in 2011, and the MSc and Ph.D. degrees from Nanyang Technological University (NTU), in 2012 and 2016, respectively. Currently, he is a professor at the University of Electronic Science and Technology of China. He served as the TPC member in several conferences such as AAAI, IJCAI. He has served as a guest editor for Pattern Recognition and Neurocomputing. His current research interests include deep learning and computer vision.
\end{IEEEbiography}

\begin{IEEEbiography}[\AddPhoto{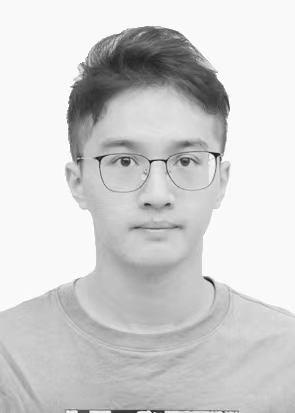}]{Haotian Lu}
is studying at the College of Artificial Intelligence, Nankai University.
He graduated from Taiyuan University of Technology with a bachelor's degree in 2022.
His research interests include computer vision and medical image processing.
\end{IEEEbiography}

\begin{IEEEbiography}[\AddPhoto{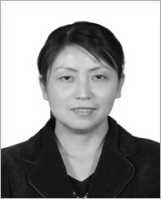}]{Jing Xu}
is a professor at the College of Artificial Intelligence, 
Nankai University. 
She received her Ph.D. degree from Nankai University in 2003.
She has published more than 100 papers in software engineering, 
software security, and big data analytics.
She won the second prize of the Tianjin Science and Technology 
Progress Award twice in 2017 and 2018, respectively.
\end{IEEEbiography}

\vfill

\end{document}